\documentclass[acmtog]{acmart}
\AtBeginDocument{%
  }

\copyrightyear{2026}
\acmYear{2026}
\setcopyright{cc}
\setcctype{by-nc-nd}
\acmConference[SA Conference Papers '26]{SIGGRAPH Asia 2026 Conference Papers}{December 01--04, 2026}{Kuala Lumpur, Malaysia}
\acmBooktitle{SIGGRAPH Asia 2026 Conference Papers (SA Conference Papers '26), December 01--04, 2026, Kuala Lumpur, Malaysia}
\acmDOI{10.1145/3829340.3842303}
\acmISBN{979-8-4007-2842-6/2026/12}

\usepackage{graphicx}
\usepackage[dvipsnames]{xcolor}
\usepackage{subcaption}
\usepackage{cleveref}
\usepackage{enumitem}
\usepackage{multirow}
\usepackage{makecell}

\begin{document}

\title{Dual-Camera Multi-Reference Smartphone Super-Resolution}

\author{Alexander Yakovenko}
\authornote{Equal contribution}
\authornote{Corresponding author}
\email{alexander.yakovenko@graphics.cs.msu.ru}
\orcid{0000-0003-3105-512X}
\affiliation{
  \institution{Huawei Noah's Ark Lab}
  \city{Shenzhen}
  \country{China}
}

\author{Aleksei Denisov}
\authornotemark[1]
\email{zeanfa@gmail.com}
\orcid{0000-0003-3105-512X}
\affiliation{
  \institution{Huawei Noah's Ark Lab}
  \city{Shenzhen}
  \country{China}
}

\author{Iaroslav Koshelev}
\authornotemark[1]
\authornote{Project lead}
\email{koshelev.iaroslav1@huawei.com}
\orcid{0000-0002-7868-2211}
\affiliation{
  \institution{Huawei Noah's Ark Lab}
  \city{Shenzhen}
  \country{China}
}

\author{Violetta Firyaridi}
\email{violettaff@gmail.com}
\orcid{0000-0002-4572-4614}
\affiliation{
  \institution{Huawei Noah's Ark Lab}
  \city{Shenzhen}
  \country{China}
}

\author{Aleksei Leonov}
\email{leonov.aleksey@h-partners.com}
\orcid{0009-0007-2295-6985}
\affiliation{
  \institution{Huawei Noah's Ark Lab}
  \city{Shenzhen}
  \country{China}
}

\author{Xueyi Zou}
\email{xy.zou@outlook.com}
\orcid{0009-0004-7565-7098}
\affiliation{
  \institution{Huawei Noah's Ark Lab}
  \city{Shenzhen}
  \country{China}
}

\renewcommand{\shortauthors}{Yakovenko, et al.}

\begin{abstract}
Modern smartphones increasingly target very high-resolution imaging\linebreak through advances in both optics and computational photography. Telephoto cameras provide access to finer spatial detail, but only cover a restricted field of view. One way to extend this detail across a larger scene is to capture multiple telephoto views and stitch them together; however, stitching lacks a global reference for preserving consistent geometry and appearance.
Conventional dual-camera reference-based super-resolution (RefSR) instead uses a wide-angle image as the reconstruction target and transfers high-resolution details from a telephoto image, but enhancement remains limited to their fixed overlap.
We generalize this setting to multi-reference super-resolution, where a wide-angle anchor is enhanced using multiple telephoto observations covering its full field-of-view. We develop a guided smartphone capture and processing pipeline for these sequentially captured, unsynchronized references, which may differ due to perspective, parallax, exposure, and occlusions. The pipeline retrieves local telephoto references, transfers their details using a RefSR transformer, and applies generative restoration conditioned on the wide-angle anchor where reliable reference information is unavailable. We demonstrate the system with scans containing up to 36 telephoto images and produce full-frame outputs at resolutions up to $15\mathrm{K}\times20\mathrm{K}$. Project page: \url{https://dual-camera-multi-ref-sr.github.io/}

\end{abstract}



\maketitle

\section{Introduction}

\begin{figure}[t]
    \centering    \includegraphics{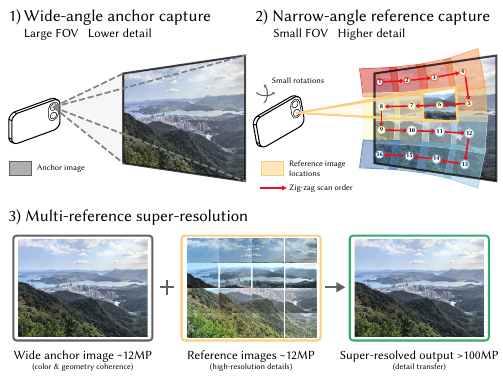}
    \caption{\textbf{Dual-camera multi-reference super-resolution.} We capture a wide-angle anchor image to define scene geometry and color, then scan it with overlapping telephoto references in a zig-zag pattern. Unlike conventional dual-camera RefSR~\cite{wang2021dual}, where telephoto detail is limited to a small central overlap, or direct stitching~\cite{brown2007automatic}, which can suffer from exposure changes, moving objects, and geometric distortions, our pipeline matches and fuses telephoto details, then refines occluded regions via generative restoration, producing a coherent ultra-high-resolution image with telephoto-level detail across the full frame.}
    \Description{Three-stage overview of the proposed capture and reconstruction process. A wide-angle anchor image first captures the full scene, multiple narrow-angle telephoto references are then acquired across its field of view following a guided zig-zag scan, and the anchor and reference images are combined to produce a full-field-of-view super-resolved output above 100MP.}
    \label{fig:teaser}
\end{figure}

\begin{figure}[t]
    \centering
    \includegraphics{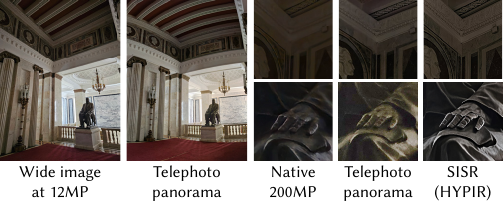}
    \caption{\textbf{Comparison of high-resolution image creation.}
Native high-megapixel capture is limited by low SNR in dim scenes, panorama stitching may distort scene geometry, and generative super-resolution can hallucinate details not supported by the input.}
    \Description{Comparison of several high-resolution imaging approaches on the same indoor scene. The figure shows a 12MP wide-angle capture, a telephoto panorama, native 200MP capture, and a telephoto panorama alongside SISR output from HYPIR, with enlarged crops illustrating low-light detail, stitching distortion, and hallucinated fine structure.}
    \label{fig:samsung_lowlight}
\end{figure}

Computational photography has expanded smartphone capabilities over the past decade, making mobile devices the primary tool for everyday photography~\cite{delbracio2021mobile}.
This evolution is driven in part by demand for higher spatial resolution, reflected in the adoption of $100+$ megapixel sensors in flagship devices~\cite{samsung100mp}.
These high-density sensors provide fine spatial detail in well-lit environments, but reduced photon collection and elevated readout noise degrade signal-to-noise ratio (SNR) under poor illumination (\cref{fig:samsung_lowlight}).
Thus, robust high-resolution, high-quality mobile photography remains an open problem.

These hardware limitations motivate algorithmic approaches, particularly single-image super-resolution (SISR), with diffusion-based priors and broader generative AI (GenAI) being increasingly integrated into mobile image signal processing (ISP) pipelines~\cite{googlegenai}.
Despite visually striking outputs, GenAI systems may hallucinate plausible but scene-inconsistent structures without genuine high-frequency information, as shown in~\cref{fig:genai_selection}.

One way to circumvent the fidelity limitations of GenAI SISR is multi-lens reference-based super-resolution (RefSR), where the missing high-frequency details are introduced by optically zoomed-in reference. 
Existing fusion of synchronized wide, ultra-wide, and telephoto views enables detail-preserving magnification beyond digital cropping~\cite{wu2023efficient}, but the highest resolution gains remain confined to the central overlap, while peripheral regions regress to SISR-like quality.

Burst photography instead fuses multiple captures of the same area and remains a standard paradigm for multi-frame SR~\cite{lecouat2022hdr} and denoising~\cite{godard2018deep}, improving effective resolution and SNR through temporal alignment.
Yet the achievable magnification is limited by lens sharpness relative to pixel size, which constrains the useful sub-pixel information available for reconstruction. In mobile systems, little additional detail is recovered beyond approximately $2\times$~\cite{wronski2019handheld}.
This limitation prevents burst SR methods from reaching $100+$ megapixel resolution.

Another route to high-resolution imagery is multi-image stitching~\cite{brown2007automatic}.
This approach increases pixel count by scanning the target field of view (FoV) with a narrower, optically zoomed-in sensor.
Compositing the resulting views yields a high-resolution output beyond the effective resolution of a wider-FoV sensor.
Despite its appeal, stitching is limited by color inconsistency, misalignment, and global geometry violations, and is typically exposed only in panorama-like capture modes.
Stitching artifacts can be mitigated by using a full-FoV wide-angle capture as a geometric and photometric anchor, a strategy aligned with RefSR.

We explore this direction with a computational photography system for dual-camera, multi-view high-resolution imaging. A wide-angle anchor frame is captured first, followed by an angularly guided scan of its FoV with the telephoto camera, producing a set of overlapping zoomed-in references (\cref{fig:teaser}). Our pipeline matches these references to the anchor and transfers their high-resolution details to local regions, producing full-frame outputs at the 100+ megapixel scale. Since the references are captured sequentially, some anchor regions may lack reliable telephoto information due to occlusions and other changes between captures. We address this by predicting where reference detail was successfully transferred and using this information to fuse the RefSR reconstruction with a GenAI-based restoration in unsupported regions.

Our contributions can be summarized as follows:
\begin{itemize}
\item\textbf{Full-FoV dual-camera RefSR.}
We extend dual-camera RefSR from a fixed wide--telephoto overlap to the full FoV of a wide-angle anchor using multiple telephoto observations.
\item \textbf{Guided acquisition and reconstruction pipeline.}
We develop a guided capture and reconstruction pipeline for unsynchronized references. It combines local reference retrieval, reference-based detail transfer, and generative restoration to handle varying reference support and scene changes.
\item \textbf{RefSR with detail-transfer mask prediction.}
We propose a RefSR transformer that jointly transfers fine details from the selected reference and predicts a detail-transfer mask. The mask lets the pipeline distinguish reference-supported from unsupported regions.
\end{itemize}

\section{Related Work}
\paragraph{Image stitching.}
A common way to create a high-resolution wide-FoV image is to combine several overlapping captures.
Early panorama systems relied on global homographies~\cite{brown2007automatic}, which work well for static scenes but can produce misalignment and ghosting in dynamic or non-planar environments.
To address this,~\cite{zaragoza2013projective} introduced spatially varying projective warps.
\cite{herrmann2018robust} used multiple registrations to better align regions at different depths.
Learning-based methods further improve parallax tolerance and visual quality, including unsupervised deep stitching~\cite{nie2023parallax}, implicit neural stitching~\cite{kim2024implicit}, diffusion-based rectangling~\cite{zhou2024recdiffusion}, and pixel-wise bidirectional warps~\cite{jin2025pixelstitch}.
Despite these advances, stitching remains sensitive to parallax, scene motion, exposure differences, and accumulated geometric distortions; it also lacks a global reference for preserving consistent geometry and appearance.

\paragraph{Burst super-resolution.}
Another line of work studies burst or multi-frame super-resolution, which fuses same-camera captures to recover detail beyond a single exposure.
These methods exploit differently aliased observations produced by sub-pixel motion to recover additional details.
\cite{wronski2019handheld} demonstrated a handheld raw burst pipeline for mobile photography that replaces demosaicing with multi-frame fusion.
Learning-based approaches further advanced this direction through deep burst reconstruction~\cite{bhat2021deepburst}, end-to-end raw burst optimization~\cite{lecouat2021lucaskande}, self-supervised training on real bursts~\cite{bhat2023self}, and joint HDR and super-resolution from raw bursts~\cite{lecouat2022hdr}.
However, because all frames are captured through the same optics, the additional high-frequency information available for reconstruction remains limited. In practice, little additional benefit is observed beyond approximately $2\times$ magnification~\cite{wronski2019handheld}.
\paragraph{Reference-based super-resolution.}
Telephoto cameras in modern smartphones provide a different source of fine detail.
Reference-based super-resolution (RefSR) can exploit such observations by matching target and reference images and transferring high-frequency details from the reference~\cite{jiang2021robust, cao2022reference, zhang2023lmr, ko2024cssr}.
However, many RefSR methods are developed and evaluated with synthetic degradations, typically bicubic downsampling, which poorly represent real sensor and ISP artifacts.

Dual-camera RefSR instead uses synchronized wide-angle and telephoto captures, addressing cross-camera degradations and geometric misalignment through supervised, self-supervised, or device-adaptive formulations~\cite{wang2021dual, zhang2022self, xu2023zero, yue2024kedusr}. Closely related, HZSR~\cite{wu2023efficient} targets hybrid-zoom enhancement while explicitly modeling occlusions and depth-of-field differences. However, in these methods telephoto detail remains confined to the fixed wide--telephoto overlap, covering only part of the wide-angle image.

Recent work combines diffusion-based generation with RefSR~\cite{guo2024refir,wang2026trust,iRAG,landmarks}, using references to guide the generation rather than explicitly transfer observed reference details. In contrast, our goal is to preserve factual reference information wherever reliable correspondences exist.
\begin{figure}[t]
    \centering
    \includegraphics[width=0.49\linewidth]{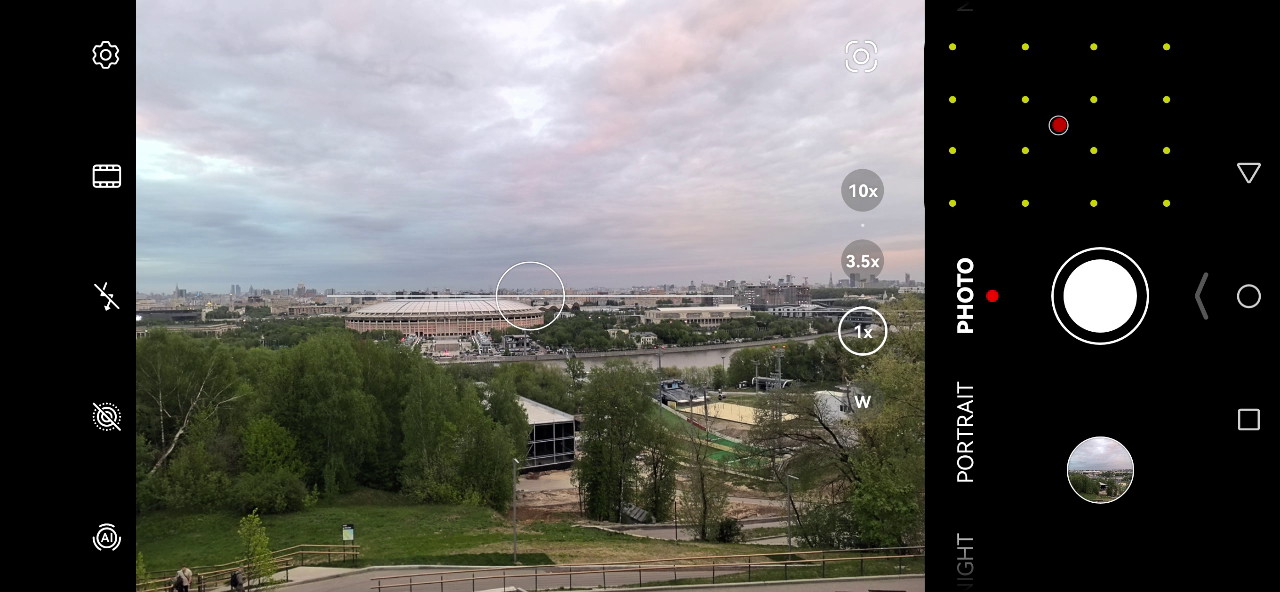}\hfill
    \includegraphics[width=0.49\linewidth]{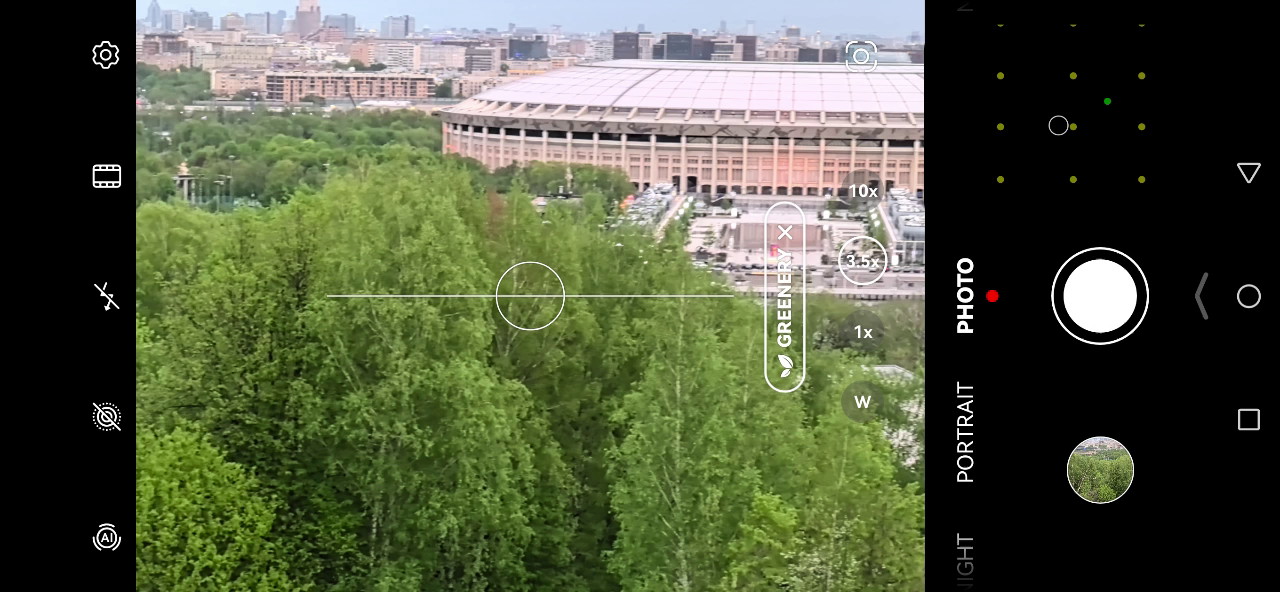}
   \caption{
    \textbf{Capture UI.}
    Picture-in-Picture guidance interface for capturing the wide-angle anchor and telephoto references, located in the top-right corner of each screenshot.
    \textbf{Left:} anchor capture; \textbf{right:} telephoto capture.
    The user aligns the target dot with the circle before each shot.
    Green/yellow dots denote anchor/reference targets; an enlarged red dot indicates correct alignment, while reduced opacity indicates incorrect roll.
    }
    \Description{Two screenshots of the smartphone capture interface with a Picture-in-Picture guidance overlay in the top-right corner. The left example shows wide-angle anchor capture and the right shows telephoto reference capture; colored target dots and an alignment indicator guide the user to the required camera orientation before each shot.}
    \label{fig:phone_ui}
\end{figure}
\paragraph{Image matching.}
With sequential wide-angle and telephoto capture, the exact location of each telephoto view within the wide-angle image is not known, requiring correspondence estimation under limited overlap and substantial scale differences.
Classical feature-based methods such as SIFT~\cite{lowe2004distinctive} and SURF~\cite{bay2008speeded} rely on sparse detected keypoints and can become unreliable in textureless or locally ambiguous regions. Detector-free learned approaches such as LoFTR~\cite{Sun_2021_CVPR} instead perform semi-dense matching directly from dense feature maps, providing correspondences beyond salient keypoints. More recent methods such as DUSt3R~\cite{dust3r_cvpr24} and MASt3R~\cite{mast3r_eccv24} further improve robustness to unconstrained viewpoint and scale changes.
However, large wide--telephoto scale differences remain challenging even for these methods.

\paragraph{Generative AI}
Although image matching and RefSR enable high-quality transfer of reference detail, this transfer is limited by visibility: occluded regions have no corresponding reference content and therefore no high-resolution information to transfer from. Generative models~\cite{goodfellow2014generativeadversarialnetworks,ho2020denoisingdiffusionprobabilisticmodels} offer a way to fill this gap by synthesizing plausible high-frequency detail conditioned on the low-resolution (LR) input.
Such conditioning helps preserve the global image structure and has been widely used for super-resolution~\cite{wang2018esrganenhancedsuperresolutiongenerative,lugmayr2020srflowlearningsuperresolutionspace,li2021srdiffsingleimagesuperresolution}.
Recent SR methods increasingly exploit diffusion priors for real-world restoration ~\cite{duan2025dit4sr, chen2025faithdiff,li2026linearsr,lin2025hypir}.
Since these methods may hallucinate details not supported by the input, their use is most appropriate in regions where sufficient reference information is unavailable.
\begin{figure}[t]
    \centering
    \begin{subfigure}[t]{0.23\linewidth}
        \includegraphics[width=\linewidth]{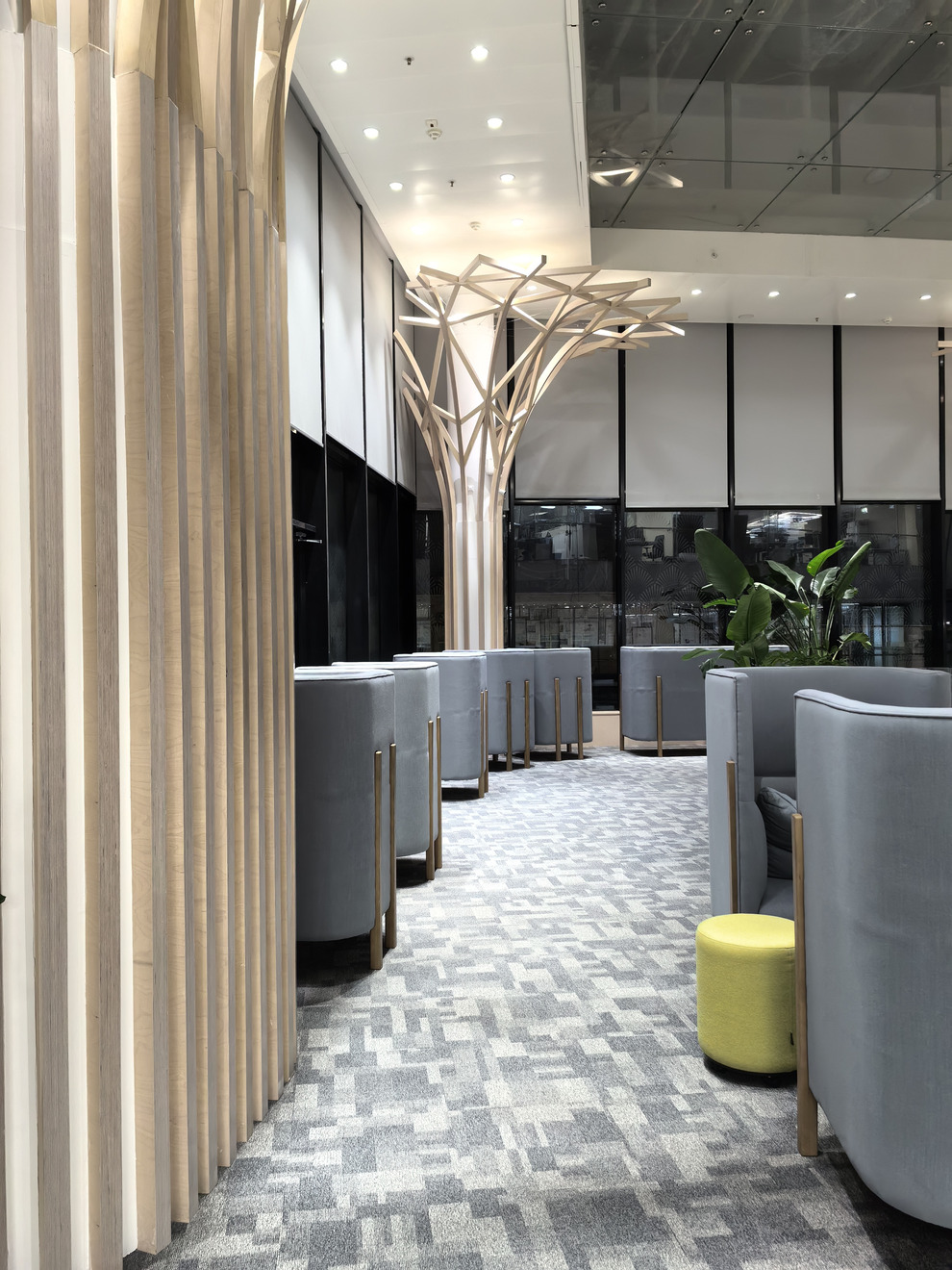}
        \caption{Anchor frame}
    \end{subfigure}
    \hfill
    \begin{subfigure}[t]{0.23\linewidth}
        \includegraphics[width=\linewidth]{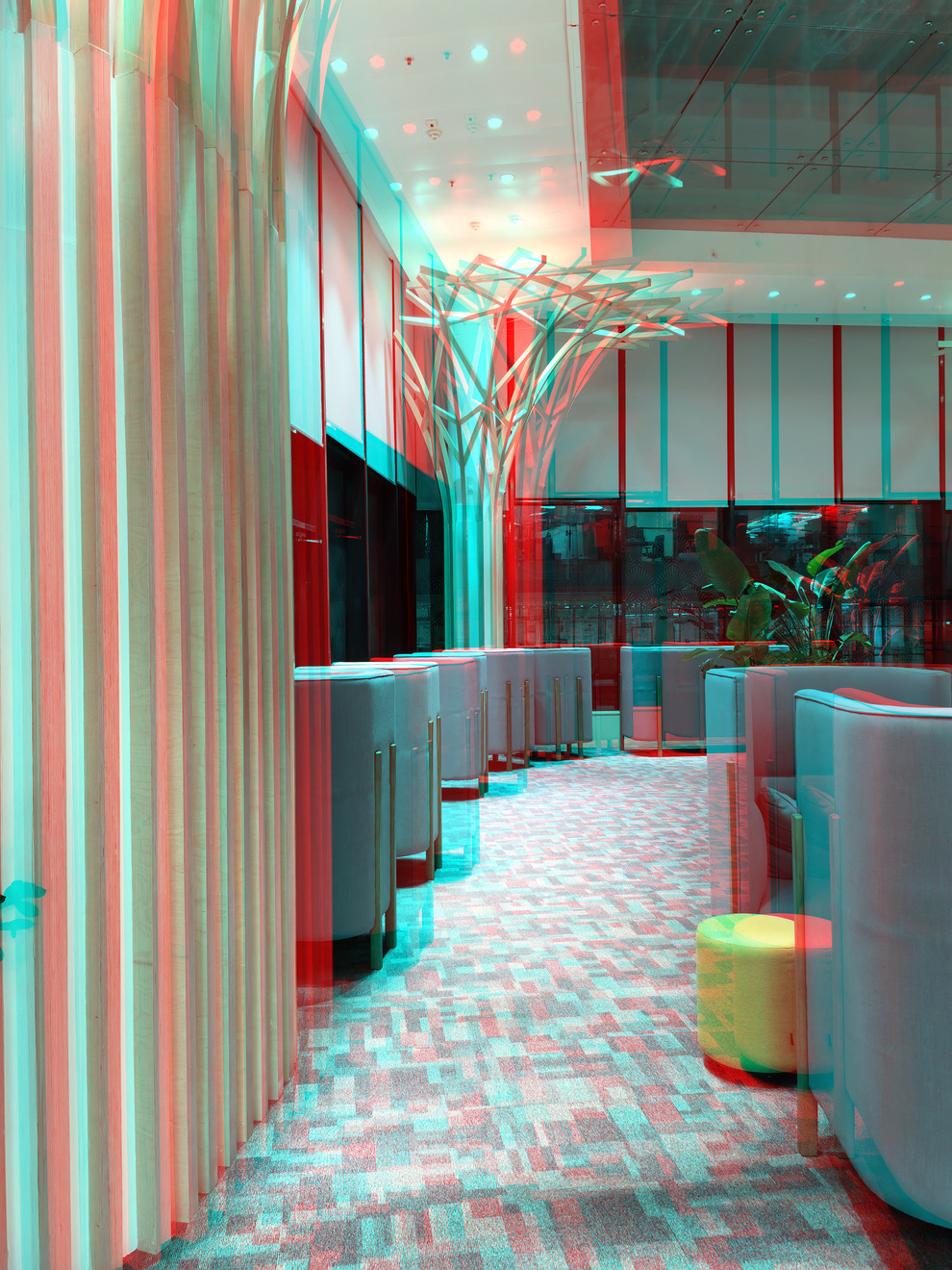}
        \caption{Anchor drift (before/after)}
    \end{subfigure}
    \hfill
    \begin{subfigure}[t]{0.23\linewidth}
        \includegraphics[width=\linewidth]{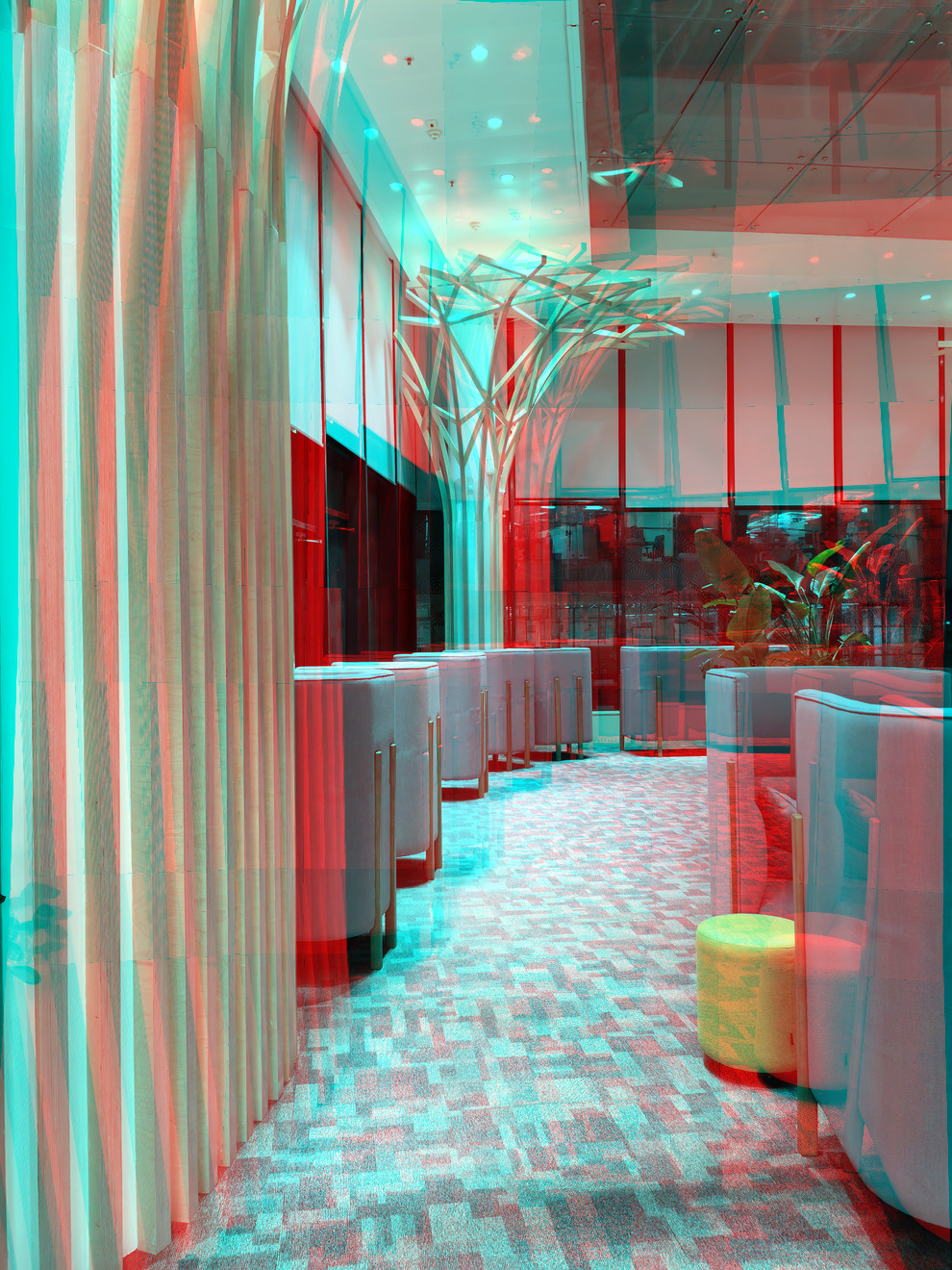}
        \caption{IMU alignment}
    \end{subfigure}
    \hfill
    \begin{subfigure}[t]{0.23\linewidth}
        \includegraphics[width=\linewidth]{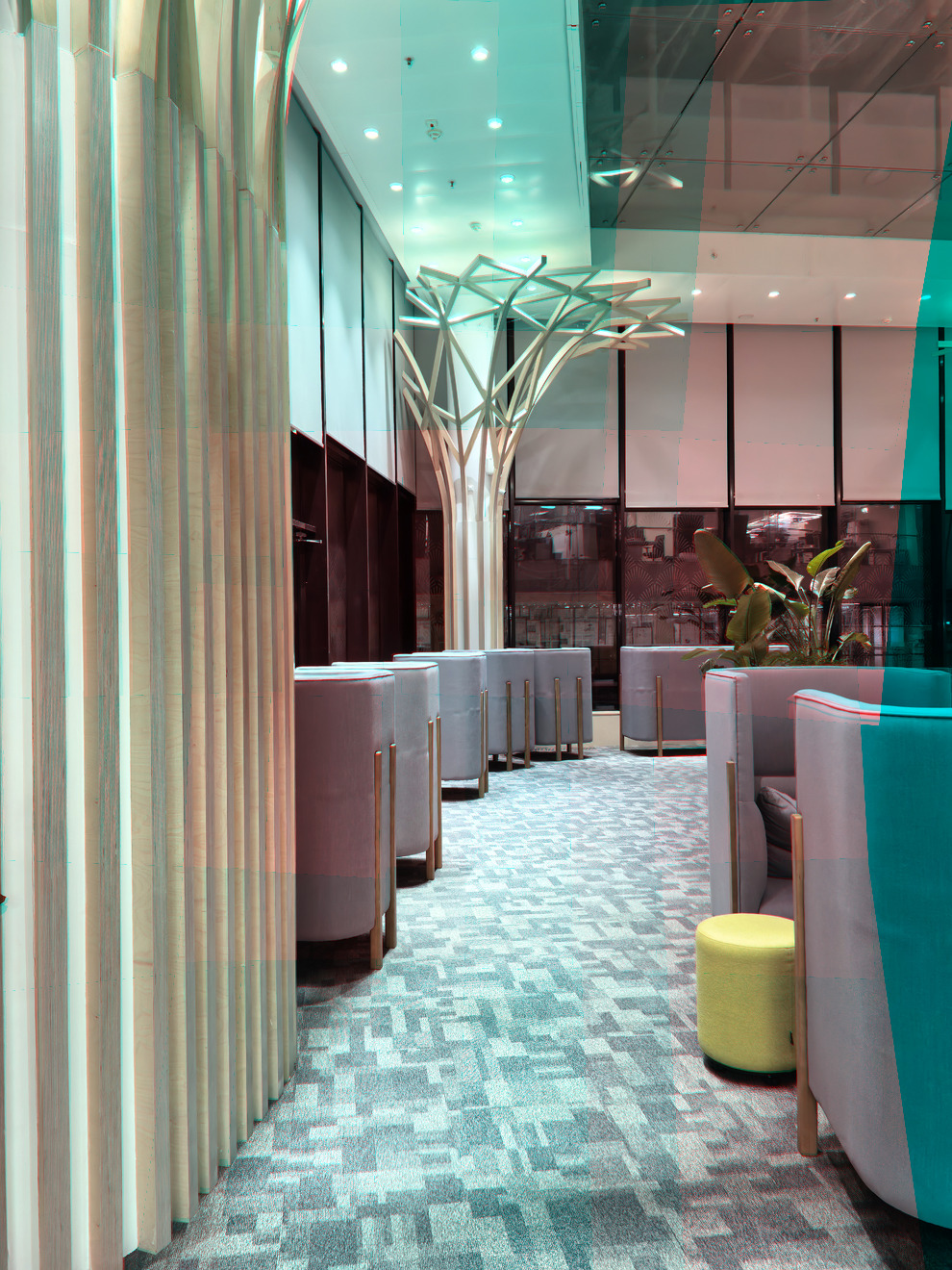}
        \caption{Homography alignment}
    \end{subfigure}
    \caption{\textbf{IMU drift and coarse matching for reference registration.} Images are overlaid as red/cyan anaglyphs; color fringing indicates misregistration. (a) Wide-angle anchor frame. (b) Before/after anchor comparison reveals IMU drift accumulated over the tele capture. (c) IMU-based projection of tele references with registration error growing as the capture progresses. (d) Coarse homography estimation substantially reduces fringing, motivating explicit feature matching over raw IMU data.}
    \Description{Four views illustrating registration error from IMU-based guidance. The figure shows the wide-angle anchor, the anchor before and after the telephoto scan, red/cyan anaglyph overlays using IMU-based alignment, and the same views after coarse homography alignment; reduced color fringing indicates improved registration.}
    \label{fig:anaglyphs}
\end{figure}

\begin{figure*}[t]
    \centering
    \includegraphics{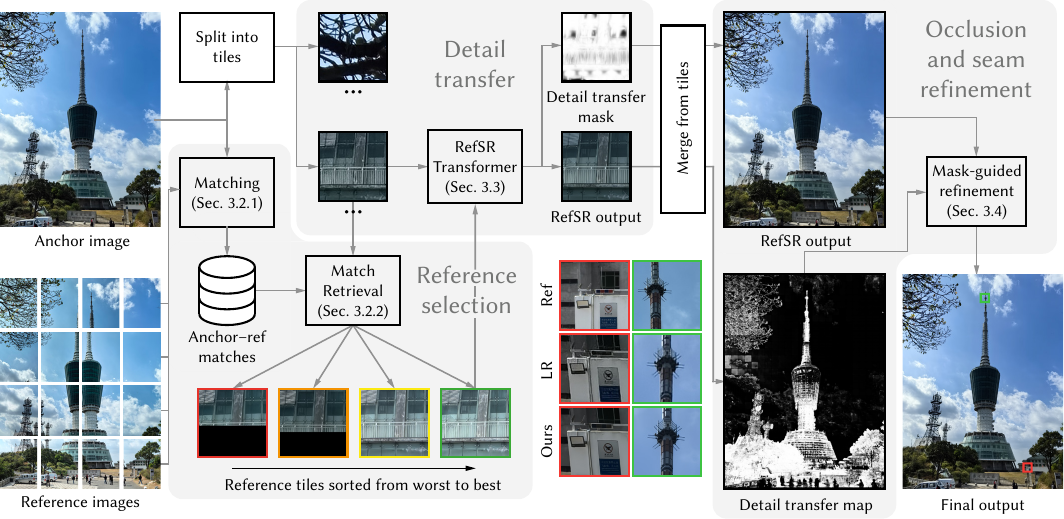}
    \caption{\textbf{Overview of the proposed reconstruction pipeline.}
    Given a wide-angle anchor image and a sequence of telephoto reference frames, we first split the anchor into overlapping tiles. For each tile, semi-dense anchor--reference matches are queried to retrieve the best telephoto crop. The anchor tile is then upsampled to the target resolution and, together with the selected reference crop, processed by the RefSR module, which predicts both a super-resolved tile and a detail-transfer mask. The resulting tiles and masks are assembled into a full-resolution RefSR image and mask. The mask is then used to preserve reliable RefSR regions and restore occluded or unsupported areas, producing the final ultra-high-resolution image.
    }
    \Description{Overview of the full reconstruction pipeline. The wide-angle anchor and telephoto reference images are first matched to establish anchor-reference correspondences. The anchor is then split into tiles, and these correspondences are queried to retrieve the best telephoto crop for each tile. RefSR transfers local detail and predicts a detail-transfer mask; the resulting tiles and masks are merged into full-resolution outputs, followed by mask-guided refinement of unsupported regions to produce the final image.}
    \label{fig:main}
\end{figure*}

\section{Dual-Camera Multi-Reference Super-Resolution}
\label{sec:method}

Our goal is to produce an ultra-high-resolution image from a dual-camera mobile capture while preserving the global appearance and geometry of a single wide-angle view. The overall acquisition and processing strategy is illustrated in \cref{fig:teaser}. We first capture a wide-angle \emph{anchor} image, which defines the scene content, field of view, color appearance, and target geometry. We then acquire a sequence of overlapping telephoto \emph{reference} images, providing high-resolution observations of different regions of the anchor frame. The captured data is then processed by our pipeline, shown in \cref{fig:main}: telephoto references are matched to the anchor image, relevant references are retrieved for each output tile, local details are transferred with a reference-based super-resolution model, and unsupported regions are restored by generative refinement.

The rest of this section describes each component of the system. We describe the capture protocol in \cref{sec:method:capturing}, anchor--reference matching and reference retrieval in \cref{sec:method:matching}, reference-based detail transfer in \cref{sec:method:refsr}, and the final unsupported region refinement and blending stage in \cref{sec:method:inpainting}.

\subsection{Image capturing setup}
\label{sec:method:capturing}

To collect the images and approximate angular information required for reconstruction, we use the device’s native camera application with overlaid angular guidance for capturing telephoto references across the wide-angle anchor FoV. We rely on native capture to obtain the best image quality available on the device, as standard Android Camera APIs~\cite{AndroidCameraAPI} do not always expose the proprietary image processing used by the camera application. The guidance is therefore implemented as a \emph{Picture-in-Picture} (PiP) overlay that uses the Android Sensor API for orientation tracking~\cite{AndroidSensorsAPI}. The user first captures a wide-angle anchor image and then telephoto references by aligning the target dot with each grid location, as illustrated in \cref{fig:phone_ui}. The references are acquired in a fixed zig-zag order starting from the top-left to reduce motion between consecutive shots (\cref{fig:teaser}).

The guidance grid is computed from two device-specific parameters: the wide-angle camera FoV and the telephoto zoom factor relative to the anchor. We estimate the telephoto FoV from the zoom factor, choose a fixed overlap ratio between neighboring telephoto views, and tile the anchor FoV with telephoto-sized angular regions at this constant overlap. The centers of these regions define the target angular locations shown in the guidance grid. We slightly enlarge the grid to compensate for projective effects and lens distortions. During capture, the current phone orientation is tracked relative to the anchor pose and used to update the dot grid in real time.

The PiP design preserves vendor image quality, but because our app runs independently from the vendor camera application, the guided angular locations remain approximate. In addition, \textit{inertial measurement unit} (IMU) orientation can drift during the telephoto scan, leading to large image-space errors at telephoto scale. \Cref{fig:anaglyphs} shows that raw angular alignment accumulates visible registration error, while image-based matching substantially improves alignment. We therefore use the target angular locations only as coarse initialization and rely on explicit anchor--reference matching for reconstruction.

\subsection{Reference matching and retrieval}
\label{sec:method:matching}

The final output resolution is determined by the telephoto zoom factor $z$. We first divide the anchor image into partially overlapping local regions at its original resolution. Each region is then upsampled by $z$ and processed independently by the RefSR module. For each region, the model requires an anchor crop and a telephoto crop observing the same scene content at the target scale. Reference matching and retrieval therefore associate each local anchor region with a suitable crop from the captured telephoto images.

\subsubsection{Image matching}
\begin{figure}
    \centering
    \begin{subfigure}[t]{0.23\linewidth}
        \includegraphics[width=\linewidth]{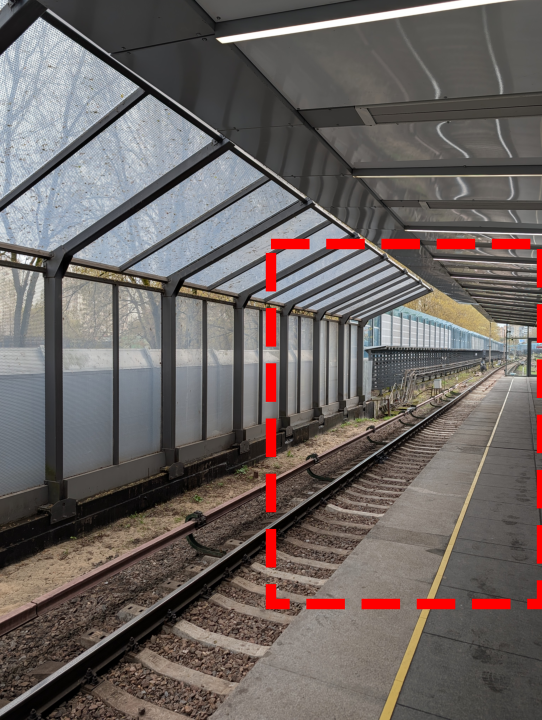}
        \caption{Anchor frame with crop region}
    \end{subfigure}
    \hfill
    \begin{subfigure}[t]{0.23\linewidth}
        \includegraphics[width=\linewidth]{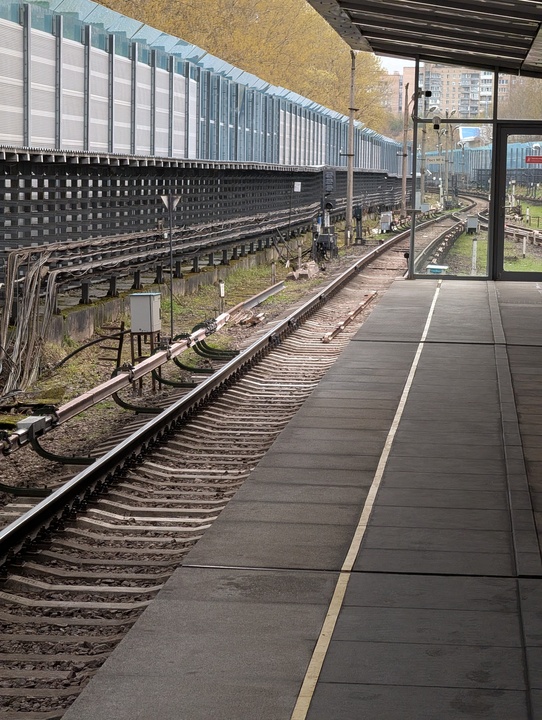}
        \caption{Reference frame}
    \end{subfigure}
    \hfill
    \begin{subfigure}[t]{0.23\linewidth}
        \includegraphics[width=\linewidth]{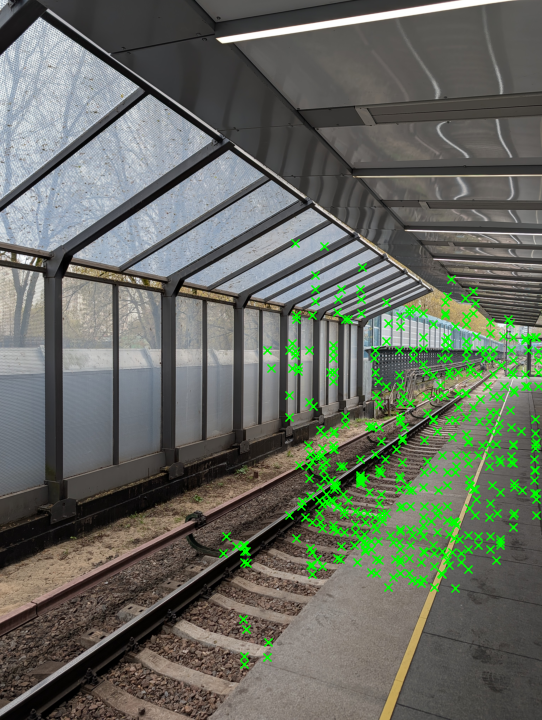}
        \caption{Matching without cropping}
    \end{subfigure}
    \hfill
    \begin{subfigure}[t]{0.23\linewidth}
        \includegraphics[width=\linewidth]{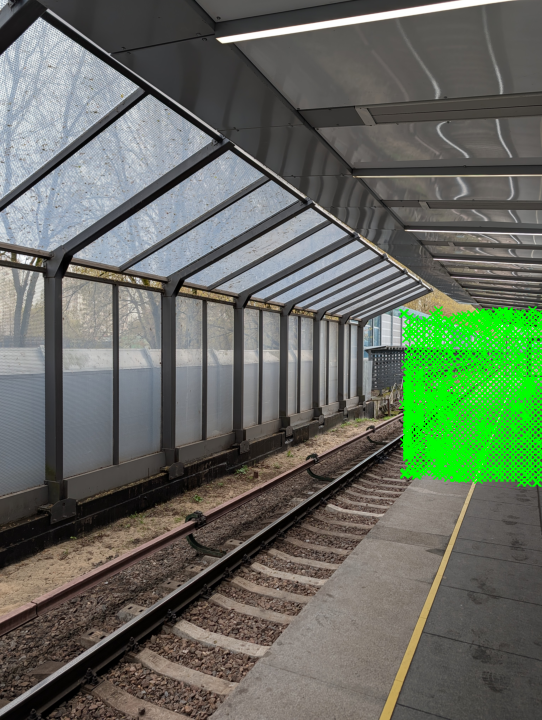}
        \caption{Matching with cropping}
    \end{subfigure}
    \caption{
        \textbf{Angularly guided image matching.}
        The angular location of each telephoto frame provides a coarse estimate of its position in the anchor image. Matching the full anchor frame to the telephoto image produces irrelevant correspondences due to the large scale difference, while matching only the angularly guided crop focuses MASt3R on the correct scene region.
    }
    \Description{Four panels show an anchor image with an angularly guided crop, the corresponding telephoto reference, matching against the full anchor, and matching within the crop. Full-image matching produces scattered and largely incorrect correspondences, while crop-based matching concentrates them in the correct scene region.}
    \label{fig:crop_matching}
\end{figure}

For each telephoto reference $I^R_i$, the capture guidance provides an approximate angular location in the anchor frame. We center a fixed-size crop at this location and use MASt3R~\cite{mast3r_eccv24} to match it against $I^R_i$, rather than matching the full anchor image. We set the crop to half the anchor width and height, which removes much of the unrelated context while retaining enough margin for pose inaccuracies. This substantially improves matching quality under large scale differences such as $5\times$ telephoto capture, as shown in \cref{fig:crop_matching}.

We map the anchor-side coordinates of the matches back to the full anchor frame and repeat this procedure for all references. This produces a correspondence set $\mathcal{M}=\{(x,y,i,x',y')\}$, where $(x,y)$ is a point in the anchor image, $i$ is the reference index, and $(x',y')$ is the corresponding point in $I^R_i$. This set stores the semi-dense\footnote{By \emph{semi-dense}, we mean correspondences estimated from dense feature maps without keypoint detection, but retained only at reliable locations rather than for every pixel.} geometric association between anchor locations and telephoto observations.

\subsubsection{Reference retrieval and filtering}

For reference retrieval, correspondences are used only to choose a telephoto crop, not to warp the reference image. This preserves the native appearance of the telephoto observation and leaves spatial adaptation to the RefSR module. For each local anchor region centered at $(X,Y)$, we query nearby matches in $\mathcal{M}$ using a KD-tree over anchor-side coordinates $(x,y)$. We use the $L_\infty$ metric with radius equal to the anchor tile size, so the query covers the square region processed by the reconstruction model.

Matches are grouped by reference index $i$. For each reference, we keep the match closest to $(X,Y)$ and use its reference-side coordinate $(x',y')$ as the candidate crop center in $I^R_i$. The reference crop is taken at the same size as the upsampled anchor region, producing at most one candidate crop per reference image.

Candidates are ranked by validity and sharpness. We first split them into valid and boundary groups using a $0.75$ valid-area threshold, preferring valid crops while retaining boundary crops as fallbacks. 
Within the selected group, sharpness is estimated as the amount of high-frequency information lost under repeated $2\times$ down- / up- sampling of the grayscale crop, with higher scores indicating sharper crops. An example of retrieval and ranking is given in \cref{fig:main}

\subsection{Reference-based super-resolution with detail-transfer masks}
\label{sec:method:refsr}

\begin{figure}
    \centering
    \includegraphics{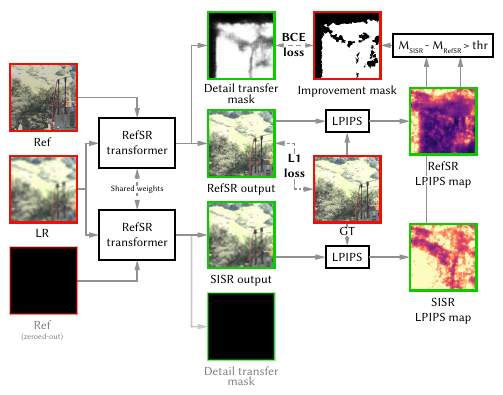}
    \caption{\textbf{Training detail-transfer mask supervision.}
During training, the RefSR transformer is evaluated twice: once with the reference crop and once without a reference, using an all-zero reference image as input. The reference-free branch is run under \texttt{no-grad} and acts as a SISR baseline. We compare spatial LPIPS maps of both outputs against the ground truth and mark regions where RefSR improves over SISR by more than a threshold as the target detail-transfer mask. The model is trained with an $L_1$ reconstruction loss for the RefSR output and a BCE loss for the predicted mask.}
    \Description{The training diagram shows two passes through the same RefSR transformer. One pass uses the reference crop and produces the RefSR output and detail-transfer mask, while the second uses an all-zero reference and serves as an SISR baseline. LPIPS maps of both outputs against the ground truth are compared to create the target mask, and the RefSR output and predicted mask are supervised with reconstruction and binary cross-entropy losses.}
    \label{fig:method:refsr_train}
\end{figure}

\begin{figure}
   
{
    \setlength{\tabcolsep}{2pt}
    \begin{tabular}{cccc}
    \includegraphics[width=0.23\linewidth,height=0.23\linewidth]{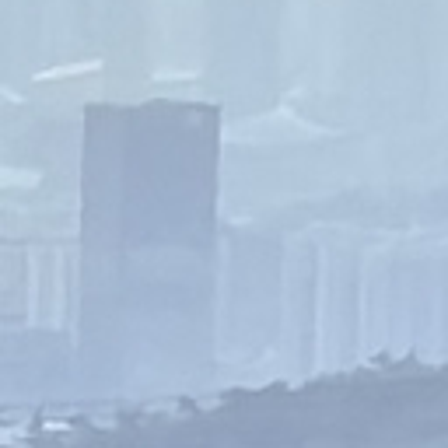} &
      \includegraphics[width=0.23\linewidth,height=0.23\linewidth]{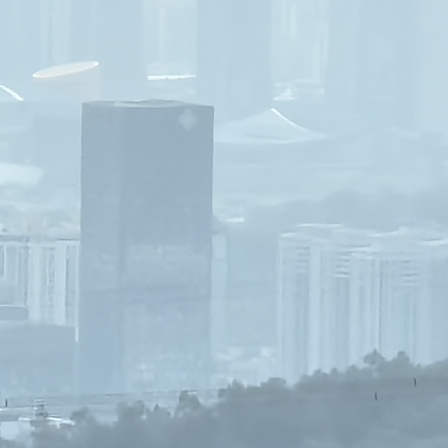} &
      \includegraphics[width=0.23\linewidth,height=0.23\linewidth]{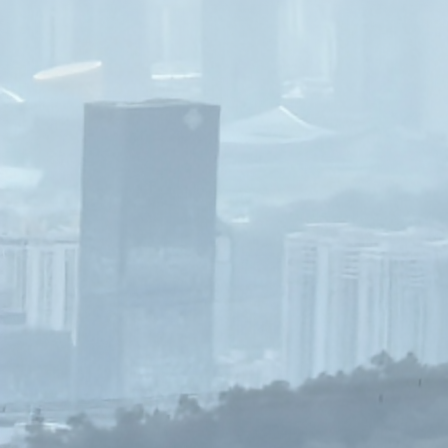} &
      \includegraphics[width=0.23\linewidth,height=0.23\linewidth]{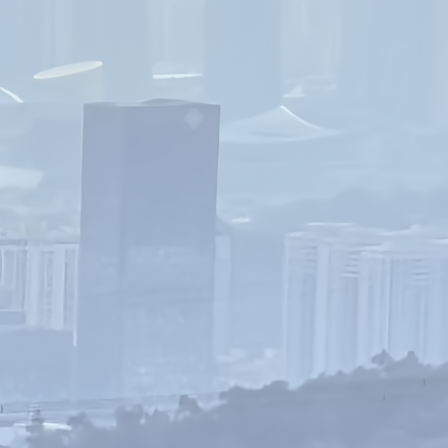} \\
      \includegraphics[width=0.23\linewidth,height=0.23\linewidth]{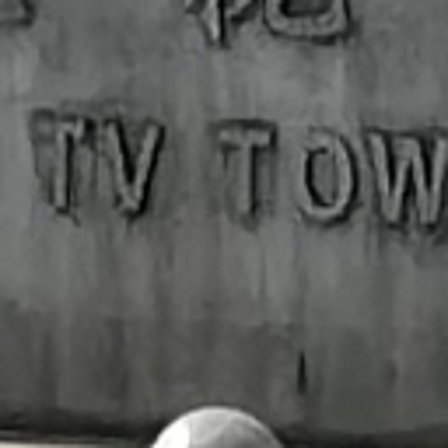} &
      \includegraphics[width=0.23\linewidth,height=0.23\linewidth]{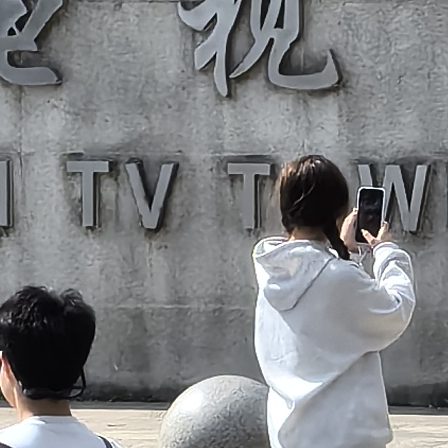} &
      \includegraphics[width=0.23\linewidth,height=0.23\linewidth]{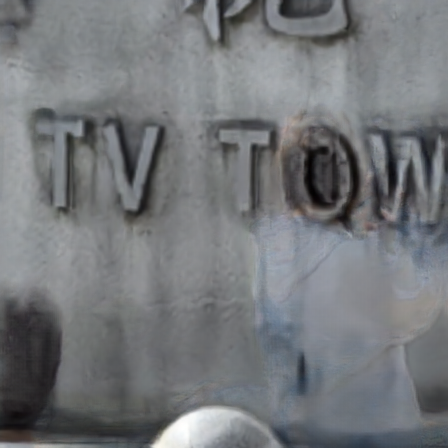} &
      \includegraphics[width=0.23\linewidth,height=0.23\linewidth]{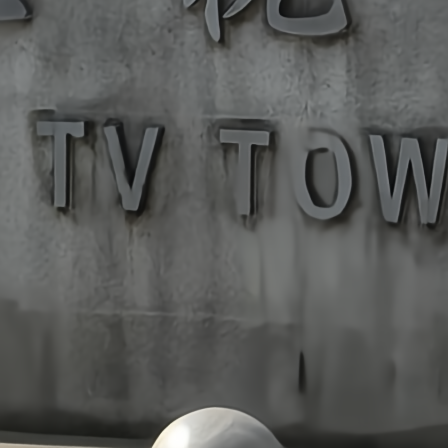}
      \\
    
      LR & Ref & KeDuSR & Ours
    \end{tabular}
}
    \caption{\textbf{Unsynchronized RefSR challenges.} Sequential capture can violate the assumptions of synchronized dual-camera methods such as KeDuSR. \textbf{Top:} In static regions, detail transfer succeeds, leaving only color mismatch. \textbf{Bottom:} When scene content changes, over-reliance on the reference can introduce ghosting and incorrect structures. In contrast, our method transfers available telephoto detail while preserving the anchor appearance.}
    \Description{Two examples compare LR input, telephoto reference, KeDuSR, and our result under unsynchronized dual-camera capture. In the top row, minor reference differences lead mainly to color mismatch, while in the bottom row, larger scene changes cause ghosting and incorrect structures in KeDuSR; our method preserves the anchor appearance while transferring available telephoto detail.}
    \label{fig:method:refsr:dual_cam_ghosting}
\end{figure}

Conventional dual-camera RefSR methods usually assume that wide and telephoto frames are captured simultaneously, so their differences are mostly limited to parallax, exposure, and ISP processing. In our capture setup, the anchor and reference images are acquired sequentially, with a time gap ranging from a few seconds to nearly a minute. During this interval, scene content may change: objects can move, become occluded, or change appearance between captures. Thus, the retrieved reference is only locally reliable, and using it uniformly can introduce ghosting or incorrect details, as shown in \cref{fig:method:refsr:dual_cam_ghosting}.

We therefore require the RefSR module to solve two coupled tasks: transfer high-frequency details where the reference is useful, and predict where such transfer actually happened. 
The \emph{detail-transfer mask} is later used to identify unsupported regions for generative refinement.
Importantly, this mask should reflect successful texture transfer rather than only geometric consistency. 
Forward--backward optical flow checks~\cite{alvarez2007symmetrical} may indicate geometric consistency even when scene changes make the corresponding reference content inconsistent with the anchor, while uncertainty predictions~\cite{ning2021uncertainty} can remain low because the LR input already provides a reasonable fallback.
To address this, we derive mask supervision from a training-time comparison between reference-based and reference-free predictions, as illustrated in \cref{fig:method:refsr_train}.

\subsubsection{RefSR transformer and training setup}

For each local reconstruction region, the model receives an LR anchor crop and a reference crop. We use a Vision Transformer that patchifies both inputs and processes the concatenated LR--reference token sequence. This lets the model learn implicit matching and texture transfer through self-attention, without explicit warping or blending. The network predicts a super-resolved RGB crop and an additional detail-transfer mask channel.

Let $I_{\mathrm{LR}}$ be the degraded crop, $I_{\mathrm{Ref}}$ the reference crop, and $I_{\mathrm{GT}}$ the ground-truth HR crop. During training, we perform two forward passes through the same network: one \emph{with the reference} and one \emph{without a reference}, implemented by replacing the reference image with an all-zero image of the same size,
\begin{equation}
\hat{I}_{\mathrm{RefSR}}, \hat{M}=f_{\theta}(I_{\mathrm{LR}}, I_{\mathrm{Ref}}),
\qquad
\hat{I}_{\mathrm{SISR}}=f_{\theta}(I_{\mathrm{LR}}, \mathbf{0}).
\end{equation}
The reference-free branch is evaluated under \texttt{no-grad} and serves as a SISR proxy, used only to estimate where the reference improves reconstruction quality.

\subsubsection{Detail-transfer mask supervision}

We compare both outputs to the ground truth using spatial LPIPS maps~\cite{lpips},
\begin{equation}
M_{\mathrm{RefSR}}=\mathrm{LPIPS}(\hat{I}_{\mathrm{RefSR}}, I_{\mathrm{GT}}),
\quad
M_{\mathrm{SISR}}=\mathrm{LPIPS}(\hat{I}_{\mathrm{SISR}}, I_{\mathrm{GT}}).
\end{equation}
Pixels where the RefSR output improves over the reference-free output by more than $\tau$ are treated as successful detail-transfer regions,
\begin{equation}
M_{\mathrm{GT}}=\left[M_{\mathrm{SISR}} - M_{\mathrm{RefSR}} > \tau\right],
\qquad
\tau = 0.1.
\end{equation}
Since both predictions are produced by the same model, this mask reflects the model's own ability to exploit the reference rather than relying on an external geometric proxy or absolute reconstruction error. This distinction is important for unsynchronized RefSR, where geometric consistency or low reconstruction error does not necessarily imply successful detail transfer when scene content changes between captures.

The RGB output is supervised with an $L_1$ reconstruction loss, while the predicted detail-transfer mask is supervised with binary cross-entropy,
\begin{equation}
\mathcal{L}_{rec}=||\hat{I}_{\mathrm{RefSR}} - I_{\mathrm{GT}}||_1,
\qquad
\mathcal{L}_{mask}=\mathrm{BCE}(\hat{M}, M_{\mathrm{GT}}).
\end{equation}
The final objective is
\begin{equation}
\mathcal{L}=\mathcal{L}_{rec}+0.01\cdot\mathcal{L}_{mask}.
\end{equation}

Together, these losses train the model to reconstruct the high-resolution output and predict where reference detail was successfully transferred.

\subsubsection{Implementation details}

Our RefSR transformer operates on $8\times8$ patches and has 177M parameters, with 24 layers and hidden dimension of 768. It is trained on 1M crop pairs extracted from 4K videos. For each pair, one video frame serves as the HR target and is degraded with a Real-ESRGAN$\times4$~~\cite{wang2021real} model to obtain the LR input, while a nearby frame is used as the reference crop, introducing natural local misalignment and appearance changes.

After processing all anchor tiles, we assemble the resulting RefSR outputs and detail-transfer masks into a full-resolution RefSR image and mask, which are then passed to the generative refinement stage.

\subsection{Generative refinement of unsupported regions}
\label{sec:method:inpainting}

\begin{figure}
    \centering
    {\includegraphics[width=\linewidth]{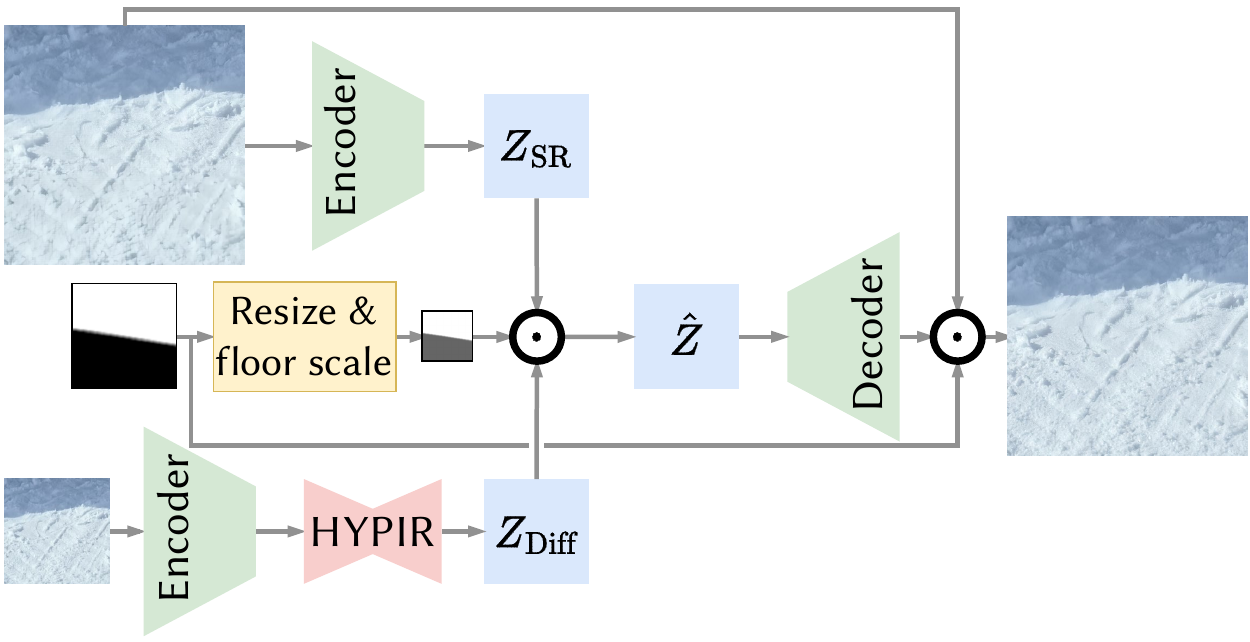}}
    \caption{\textbf{Generative refinement module.} We adapt HYPIR~\cite{lin2025hypir} for mask-guided generative refinement of unsupported regions. Reference-supported regions are preserved from the RefSR result, while unsupported ones are refined by the generative restoration model. The mask denotes supported regions in white and unsupported regions in black.}
    \Description{Diagram of the generative refinement stage. The RefSR result and upsampled LR image are encoded into latent representations, with HYPIR producing a generative restoration latent from the LR input. A resized detail-transfer mask blends the RefSR and generative latents before decoding, and the original mask is used again to combine the decoded result with the RefSR image.}
    \label{fig:inpainting}
\end{figure}

The RefSR stage transfers details only where a reliable telephoto reference is available. Regions affected by occlusions, disocclusions, scene motion, or local matching failures should instead be restored without forcing inconsistent reference texture. We therefore use the predicted detail-transfer mask to preserve reliable RefSR regions and apply generative restoration only to unsupported areas, as illustrated in \cref{fig:inpainting}.

\begin{figure}[t]
    \centering
    \setlength{\tabcolsep}{1pt}
    \begin{tabular}{cccccc}
        \includegraphics[width=0.19\linewidth]{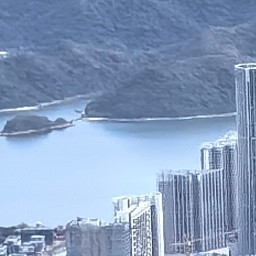} &
        \includegraphics[width=0.19\linewidth]{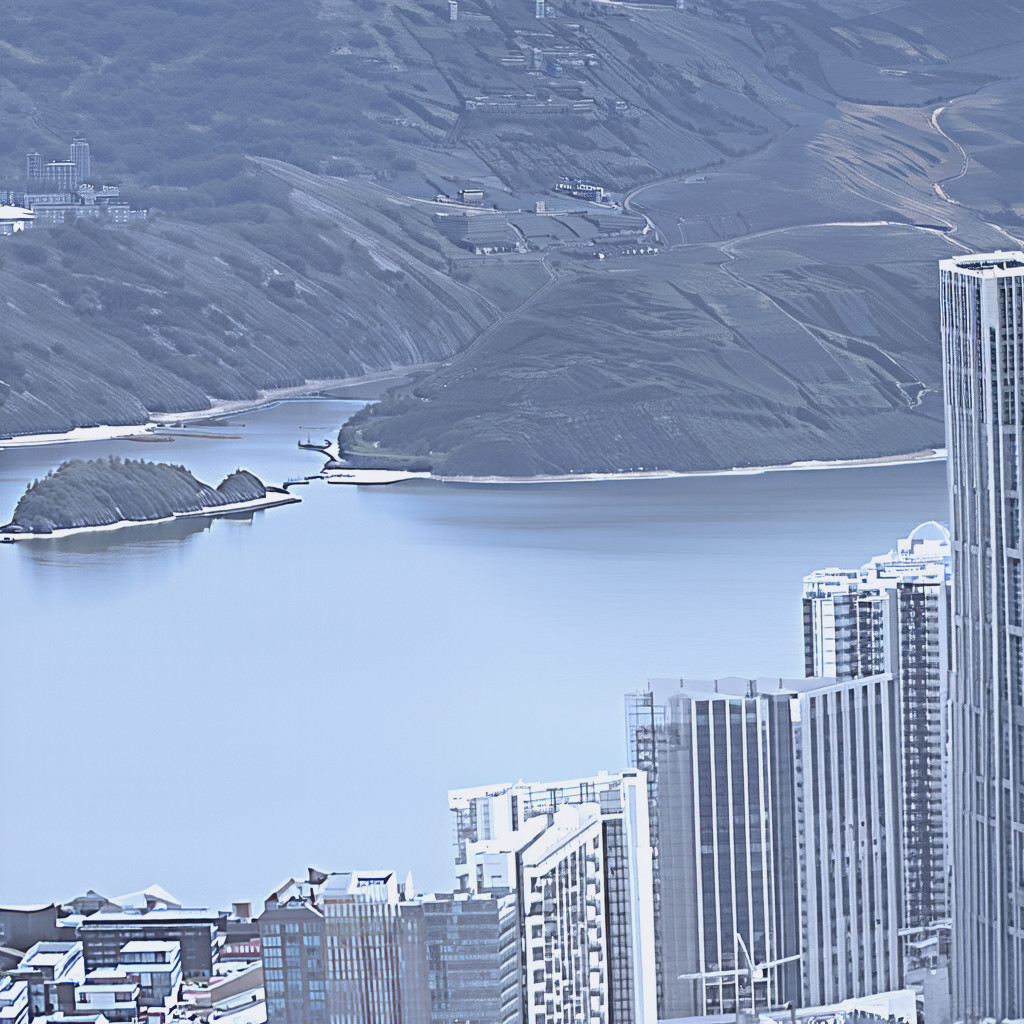} &
        \includegraphics[width=0.19\linewidth]{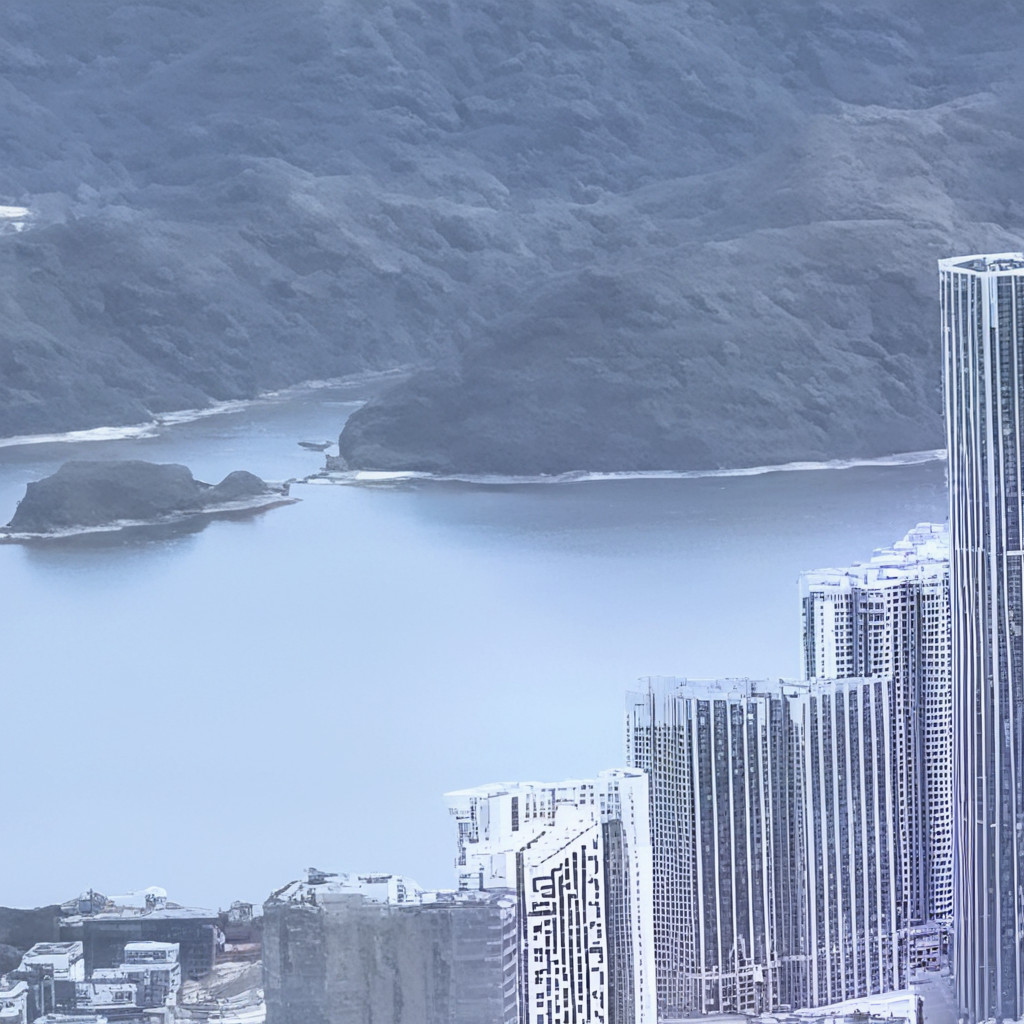} &
        \includegraphics[width=0.19\linewidth]{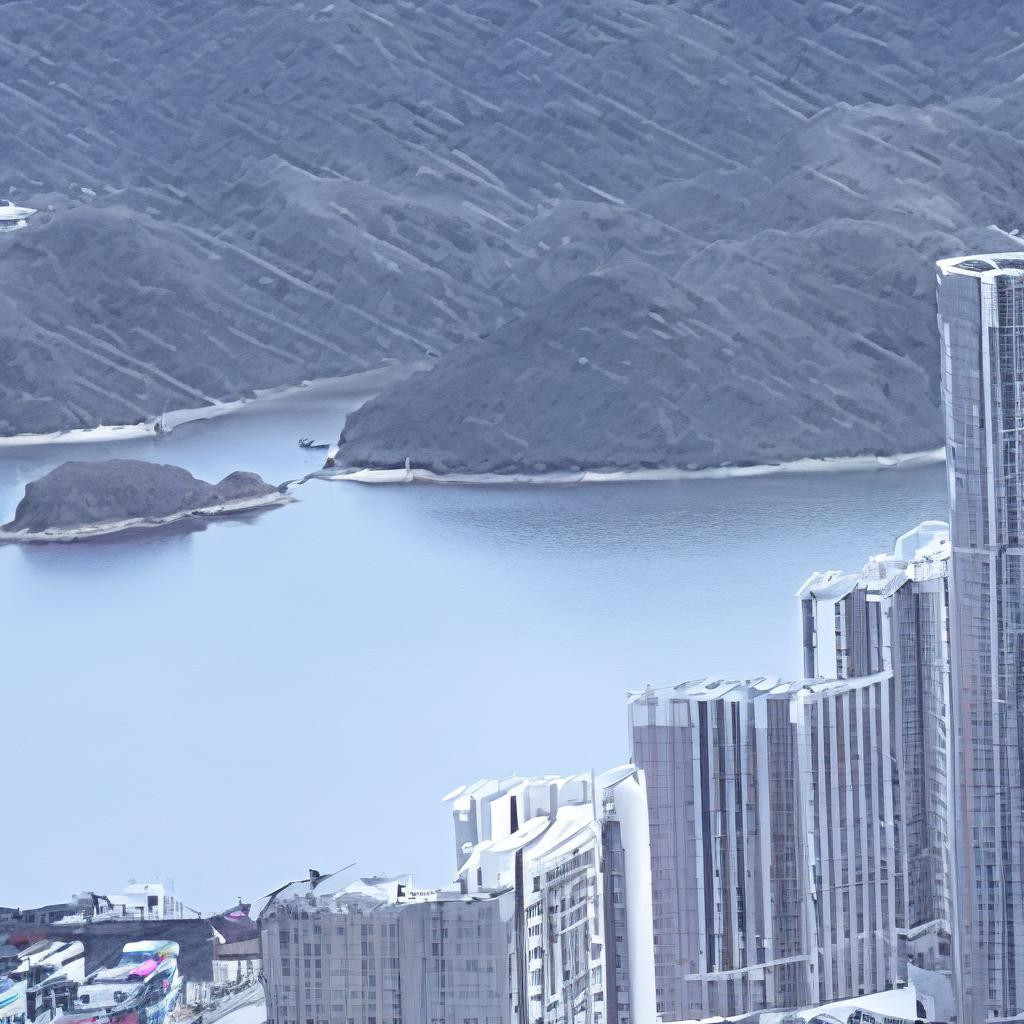} &
        \includegraphics[width=0.19\linewidth]{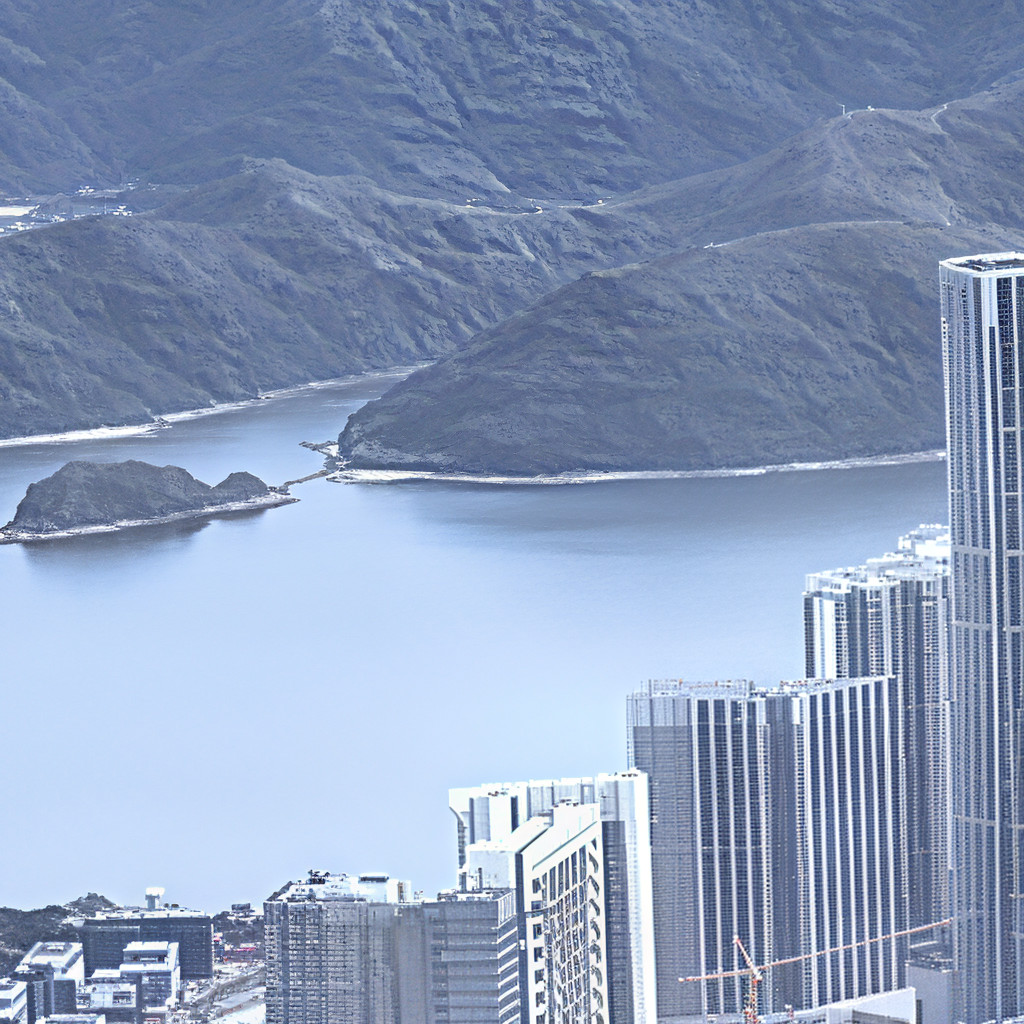} \\
        
        \small Input &
        \small DiT4SR, 45s &
        \small FaithDiff, 3s &
        \small LinearSR, 1.5s &
        \small \textbf{HYPIR, 0.5s}
    \end{tabular}
    \caption{\textbf{Selection of generative backbone.} Qualitative comparison of diffusion-based restoration methods. HYPIR provides the best quality--runtime tradeoff for our setting. Runtime is reported per $1024\!\times\!1024$ crop.}
    \Description{Qualitative comparison of four diffusion-based restoration backbones on the same input crop. The outputs of DiT4SR, FaithDiff, LinearSR, and HYPIR are shown with their per-crop runtimes, illustrating the quality–runtime tradeoff used to select HYPIR.}
    \label{fig:genai_selection}
\end{figure}

We select HYPIR~\cite{lin2025hypir}, a single-image generative restoration model, as the backbone for refining unsupported regions.
Compared to the other existing diffusion-based SISR models, as we qualitatively demonstrate in \cref{fig:genai_selection}, HYPIR provides a strong prior with efficient single-step inference. We use it with an empty prompt and do not employ external captioning, avoiding additional semantic bias and computational overhead.

Let $\hat{I}_{\mathrm{RefSR}}$ be the assembled RefSR image, $I_{\mathrm{LR}}$ the upsampled anchor image, and $\hat{M}$ the reliability mask, where $\hat{M}\!=\!1$ denotes regions preserved from RefSR and $\hat{M}\!=\!0$ denotes unsupported regions. For latent-space blending, we linearly rescale the mask values from $[0,1]$ to $[m_{\min},1]$, with $m_{\min}\!=\!0.5$ in our experiments. The rescaled mask $\widetilde{M}$ is used only in latent space, while the original mask $\hat{M}$ is kept for final pixel-space fusion.

We encode  $I_{\mathrm{LR}}$ and $\hat{I}_{\mathrm{RefSR}}$ with the HYPIR VAE. HYPIR processes the LR latent to produce the generative restoration latent $Z_{\mathrm{Diff}}$, while the RefSR image gives the latent $Z_{\mathrm{RefSR}}$. The rescaled mask is downsampled to latent resolution, yielding $\widetilde{M}_z$, and the two latents are blended as
\begin{equation}
\hat{Z}=(1-\widetilde{M}_{z}) \odot Z_{\mathrm{Diff}}+\widetilde{M}_{z} \odot Z_{\mathrm{RefSR}} .
\end{equation}
Since $\widetilde{M}_z \geq m_{\min}$, the RefSR latent remains present even inside unsupported regions. 
This constrains the generative restoration, helping suppress hallucinations and artifacts at transitions between supported and unsupported regions.

After decoding $\hat{Z}$ into $\hat{I}_{\mathrm{dec}}$, we perform a final pixel-space fusion with the original, unscaled mask:
\begin{equation}
I_{\mathrm{out}} = (1-\hat{M}) \odot \hat{I}_{\mathrm{dec}} + \hat{M} \odot \hat{I}_{\mathrm{RefSR}} .
\end{equation}

This second blend preserves reliable RefSR regions exactly, avoids VAE encode--decode degradation where generative refinement is unnecessary, and produces the final output of our pipeline.

\section{Experiments}
In this section, we evaluate the full pipeline through visual comparisons with high-resolution imaging baselines on captures from several devices, quantify the performance of our RefSR stage, discuss key design choices and computational requirements, and analyze the limitations of the proposed system. Full-resolution outputs for visual comparison are provided in the supplementary materials.

\subsection{Full-pipeline visual comparison}

In order to evaluate our full pipeline, we capture and reconstruct a range of scenes with varying content and illumination. We used five devices of different vendors and with different telephoto zoom factors ranging from $3\times$ to $5\times$.  
\paragraph{Setup} Specifically, we use the Samsung Galaxy S23 Ultra with $3\times$ zoom; Huawei P50 Pro, Mate 60 Pro+, and Pura 70 Pro with $3.5\times$ zoom; and Google Pixel 10 Pro with $5\times$ zoom.
During capture, we focus on common use cases, ranging from relatively simple static landscapes and group photos with moderate motion to challenging low-light and highly dynamic scenes.

We compare our results against four complementary baselines: native 200MP capture where available, HYPIR~\cite{lin2025hypir} as the SISR baseline, KeDuSR~\cite{yue2024kedusr} applied after our matching stage as the RefSR baseline, and Adobe Photoshop Photomerge~\cite{photomerge} as the stitching baseline.

All wide-angle anchors are captured at 12MP, resulting in final outputs from 108MP to 300MP depending on the telephoto zoom factor. Since ground-truth images at these resolutions are unavailable for real captures, we rely on full-resolution visual comparison.

\paragraph{Qualitative comparison} \Cref{fig:fig11} highlights the main qualitative differences across several representative scenarios, organized by rows as follows:
\begin{enumerate}
    \item \emph{Exposure and color changes.} KeDuSR retains the reference colors, while Photomerge only partially compensates for exposure differences between neighboring views. Our reconstruction instead preserves the anchor appearance.
    \item \emph{Local motion.} Small changes between anchor and reference capture lead to visible ghosting in KeDuSR.
    \item \emph{Text and focus variations.} HYPIR hallucinates character-like structures, while Photomerge exhibits abrupt sharpness transitions between differently focused references.
    \item \emph{Static high-detail scene.} Photomerge introduces substantial geometric distortion, while our result follows the anchor geometry. In this example, despite its lower resolution, our 108MP reconstruction has finer detail than the native 200MP capture.
    \item \emph{Dynamic scenes.} \textbf{Left:} Changing illumination introduces visible color discontinuities between neighboring telephoto views in Photomerge. Our method still transfers telephoto detail while preserving the anchor appearance. \textbf{Right:} Large scene motion makes the telephoto views mutually inconsistent, so they can no longer be stitched into a single geometrically consistent image. Our method suppresses unreliable detail transfer in the affected regions and falls back to generative refinement.
\end{enumerate}
Overall, stitching provides a strong reference point in static scenes because it uses the same telephoto observations and can retain comparable local detail. Our main advantage is that the wide-angle anchor fixes the output geometry and appearance, while the detail-transfer mask allows the reconstruction to reduce reliance on telephoto observations when they become inconsistent or unavailable.
\paragraph{Full-resolution reconstruction.}
\Cref{fig:fullimage_trimmed} shows a complete 300MP output of our full pipeline from a Google Pixel 10 Pro capture. The reconstruction uses a 12MP wide-angle anchor and 36 telephoto references at $5\times$ zoom, producing a full-frame $15\mathrm{K}\times20\mathrm{K}$ result. The crops compare the corresponding LR regions with our reconstruction and show the recovered fine detail across the FoV.

\begin{table}[t]
    \caption{\textbf{Quantitative evaluation on real dual-camera $\times2$ benchmarks.} We compare our RefSR transformer with prior dual-camera RefSR methods on synchronized wide--telephoto captures. Our model consistently achieves the strongest reconstruction quality across the evaluated datasets.}
    \centering
    \resizebox{\linewidth}{!}{
        \setlength{\tabcolsep}{3pt}
\begin{tabular}{lcccccc}
    \toprule
\multirow{2}[1]{*}{Method} & DuSR-Real & CameraFusion-Real & RealMCVSR-Real \\
    \cmidrule(lr){2-2}\cmidrule(lr){3-3}\cmidrule(lr){4-4}
     & \footnotesize{PSNR$\uparrow$/SSIM$\uparrow$/LPIPS$\downarrow$} & \footnotesize{PSNR$\uparrow$/SSIM$\uparrow$/LPIPS$\downarrow$} & \footnotesize{PSNR$\uparrow$/SSIM$\uparrow$/LPIPS$\downarrow$} \\
     \midrule
     DCSR\cite{wang2021dual}     & 26.77 / 0.875 / 0.134 & 26.02 / 0.812 / 0.293 & 26.28 / 0.811 / 0.217 \\
     SelfDZSR\cite{zhang2022self} & 26.27 / 0.856 / 0.158  & 25.94 / 0.804 / 0.283 & 25.33 / 0.793 / 0.246 \\
     ZeDuSR\cite{xu2023zero}   & 25.41 / 0.823 / 0.191  & 26.16 / 0.792 / 0.279 & 24.98 / 0.770 / 0.262 \\
     KeDuSR\cite{yue2024kedusr}   & 27.66 / 0.889 / 0.117  & 27.53 / 0.829 / 0.276 & 27.05 / 0.841 / 0.180 \\
     Our RefSR stage     & \textbf{27.93} / \textbf{0.891} / \textbf{0.113}  &  \textbf{28.21} /  \textbf{0.851} /  \textbf{0.229} & \textbf{27.21} / \textbf{0.891} / \textbf{0.164} \\
     \bottomrule
\end{tabular}
    }
    \label{tab:dusr_real}
\end{table}

\subsection{RefSR evaluation}

We evaluate the RefSR stage separately on real dual-camera benchmarks in the $2\times$ setting, following~\cite{yue2024kedusr}, and on CUFED5 with Real-ESRGAN$\times4$ degradation, following the ReFIR~\cite{guo2024refir} setting.
\paragraph{Real dual-camera RefSR}
As shown in \cref{tab:dusr_real}, our transformer consistently outperforms existing dual-camera RefSR methods on real synchronized captures. \Cref{fig:method:refsr:dual_cam_ghosting} further demonstrates its robustness when the reference content changes between sequential captures.
\begin{table}[t]

\caption{\textbf{Quantitative evaluation on CUFED5 under Real-ESRGAN$\times4$ degradation.} We compare our RefSR transformer with diffusion-based RefSR methods and the HYPIR SISR baseline. Our model achieves substantially higher distortion fidelity, while the generative methods obtain stronger perceptual and no-reference quality scores.}
    \centering
    \resizebox{\linewidth}{!}{
    \begin{tabular}{lcccc}
\toprule
\multirow{2}[1]{*}{Method} &
\multicolumn{4}{c}{CUFED5} \\
\cmidrule(lr){2-5}
&
PSNR$\uparrow$ &
SSIM$\uparrow$ &
LPIPS$\downarrow$ &
CLIP-IQA$\uparrow$ \\
\midrule

HYPIR~\cite{lin2025hypir}
& 19.34
& 0.503
& 0.359
& \textbf{0.5759} \\

ReFIR (SeeSR)~\cite{guo2024refir}
& 20.22
& 0.526
& 0.345
& 0.5538 \\

AdaRefSR~\cite{wang2026trust}
& 20.48
& 0.546
& \textbf{0.289}
& 0.5174 \\

Our RefSR stage
& \textbf{23.32}
& \textbf{0.678}
& 0.355
& 0.2821 \\
\bottomrule
\end{tabular}
    }
    \label{tab:diffusion-main}
\end{table}
\paragraph{Comparison with diffusion-based RefSR}
On CUFED5, our model substantially outperforms Ada-RefSR~\cite{wang2026trust} in PSNR and SSIM~\cite{ssim}, while Ada-RefSR achieves stronger LPIPS and CLIP-IQA~\cite{wang2023exploring}. This reflects the different objectives: our model prioritizes faithful transfer of observed reference detail, whereas generative RefSR favors perceptually plausible texture. Pure HYPIR further improves CLIP-IQA despite using no reference, illustrating the same tradeoff.

\subsection{Implementation and efficiency}

\paragraph{Implementation details.}
For detail transfer stage, we divide the anchor into $128\times128$ tiles with a 2-pixel overlap at the original anchor resolution. For each tile, the correspondence set is queried to retrieve the corresponding telephoto reference crop, and the tile is processed by the RefSR model at the target resolution. The resulting RefSR tiles are merged with Gaussian weights to form the full-resolution RefSR image passed to the refinement stage.

The generative refinement stage is processed tile-wise for memory efficiency. VAE encoding uses $1024$ px image-space tiles with stride $256$ px. Since HYPIR operates in an $8\times$ smaller latent space, the generator and VAE decoder use $128$ px latent tiles with stride $32$ px. Overlapping predictions are merged in latent space with Gaussian weights.
The RefSR and HYPIR latents are then blended using the detail-transfer mask, after which the blended latent is decoded tile-wise and assembled into the full-resolution output.

\paragraph{Refinement design choices.}
For latent blending, we use $m_{\min}~=~0.5$. Lower values give the generative restoration more freedom but increase unwanted hallucinations and lead to artifacts around transitions between supported and unsupported regions, while higher values increasingly suppress texture refinement. We found $0.5$ to provide the best balance in our experiments.

\begin{figure}[t]
    \centering
    \setlength{\tabcolsep}{2pt}
    \begin{tabular}{cccc}
        \includegraphics[width=0.23\linewidth,height=0.23\linewidth]{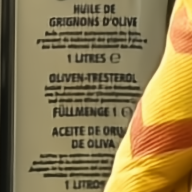} &
        \includegraphics[width=0.23\linewidth,height=0.23\linewidth]{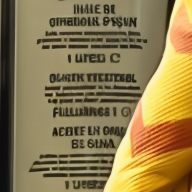} &
        \includegraphics[width=0.23\linewidth,height=0.23\linewidth]{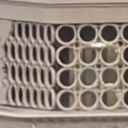} &
        \includegraphics[width=0.23\linewidth,height=0.23\linewidth]{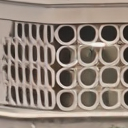} \\
        \small \makecell{RefSR output} &
        \small \makecell{VAE output} &
        \small \makecell{RefSR output} &
        \small \makecell{VAE output}
    \end{tabular}
    \caption{\textbf{VAE encode--decode degradation.}
    In regions with reliable reference transfer, passing the RefSR result through the HYPIR VAE degrades fine details. This motivates the final pixel-space blend, which preserves these regions directly from the original RefSR output.}
    \Description{Two crop pairs compare the RefSR output with the corresponding VAE output. VAE encoding and decoding alter fine semantic and geometric details: text-like structures become corrupted or unreadable, while regular shapes are visibly distorted.}
    \label{fig:vae_degradation}
\end{figure}

Passing reliable RefSR regions through the HYPIR VAE also introduces visible loss of fine detail, as shown in \cref{fig:vae_degradation}. This motivates the final pixel-space blend, which restores reliable RefSR regions directly from the original RefSR output after decoding.

\begin{table}[h]
    \caption{\textbf{Per-stage runtime of our method} on a consumer-grade GPU, reported in seconds per output megapixel.}
    \centering

\begin{tabular}{lcccc}
\toprule
& Matching & RefSR & Refinement & Total \\
\midrule
Runtime (s/MP) & 0.06 & 1.74 & 1.28 & 3.08 \\
\bottomrule
\end{tabular}
    \label{tab:runtime}
\end{table}

\subsection{Runtime}
Our design allows inference on a single consumer-grade GPU in a reasonable amount of time. We report the detailed per-stage per-megapixel timings in~\cref{tab:runtime}. Overall, for $3.5\times$ SR our method processes $4\times4$ ref grid in $\sim\!420$s, delivering a $10K \times 14K$ (140MP) image. For $5\times$ zoom factor it processes $6\times6$ grid in $\sim\!850$sec, producing a $15K\!\times\!20K$ (300MP) image.

\subsection{Limitations}
\begin{figure*}
    \centering
    \includegraphics{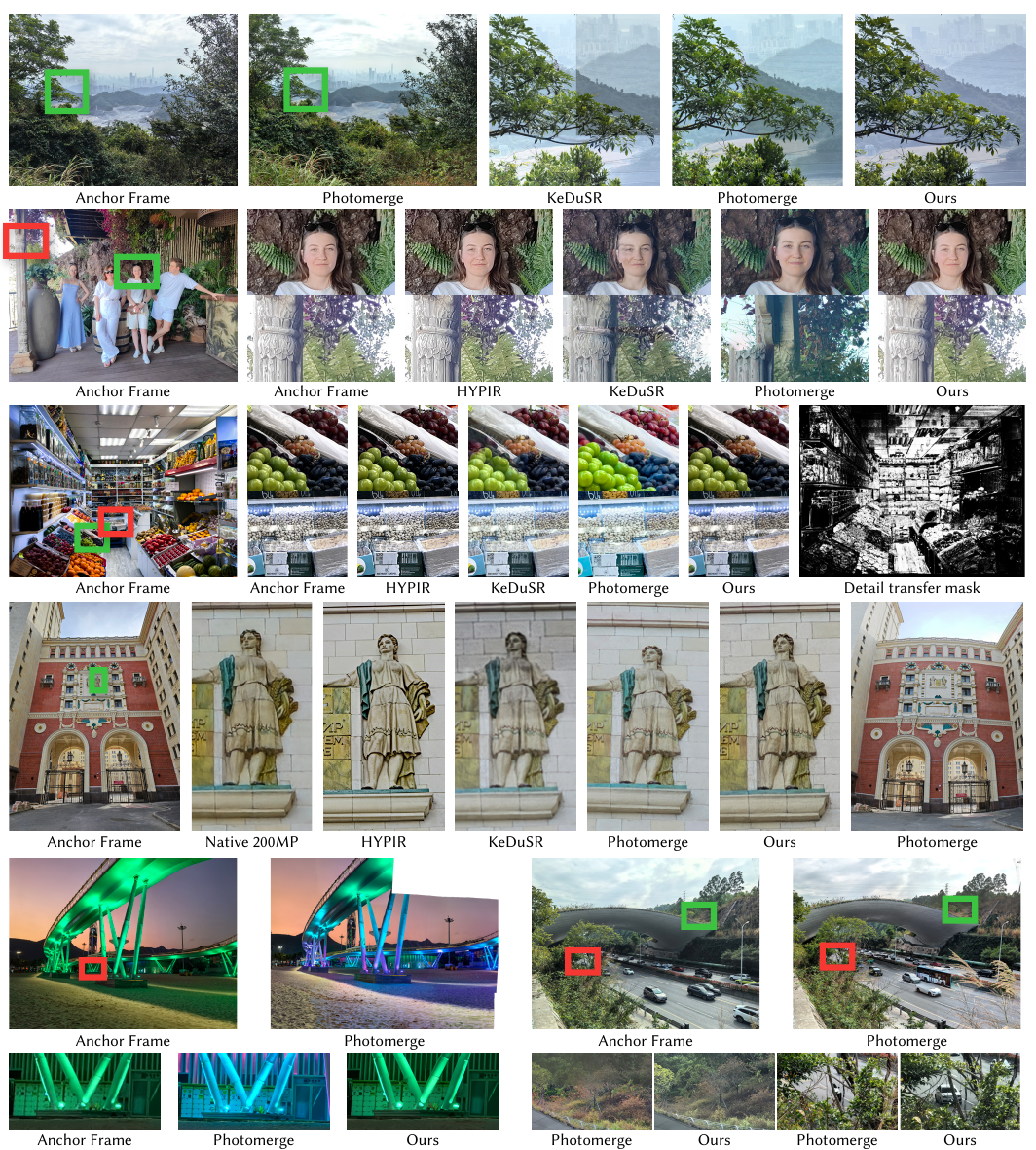}
    \caption{\textbf{Visual comparisons of our full pipeline.} Our method preserves the anchor-defined appearance and geometry while transferring fine telephoto details. KeDuSR shows ghosting under scene changes, HYPIR hallucinates fine structures, while Photomerge preserves local telephoto detail but develops exposure or focus seams and geometric distortions when neighboring views are inconsistent. White regions in the detail-transfer mask indicate reference-supported areas.}
    \Description{Qualitative comparison across several smartphone capture scenarios, including landscape, group portrait, supermarket, architectural, low-light, and dynamic traffic scenes. Crops compare the anchor input, HYPIR, KeDuSR, Photomerge, native high-resolution capture where available, and our method, illustrating baseline failures such as color inconsistency, ghosting, hallucinated structures, focus discontinuities, and geometric or stitching artifacts; one example also shows the predicted detail-transfer mask.}
    \label{fig:fig11}
\end{figure*}

\begin{figure*}
    \centering
    \includegraphics{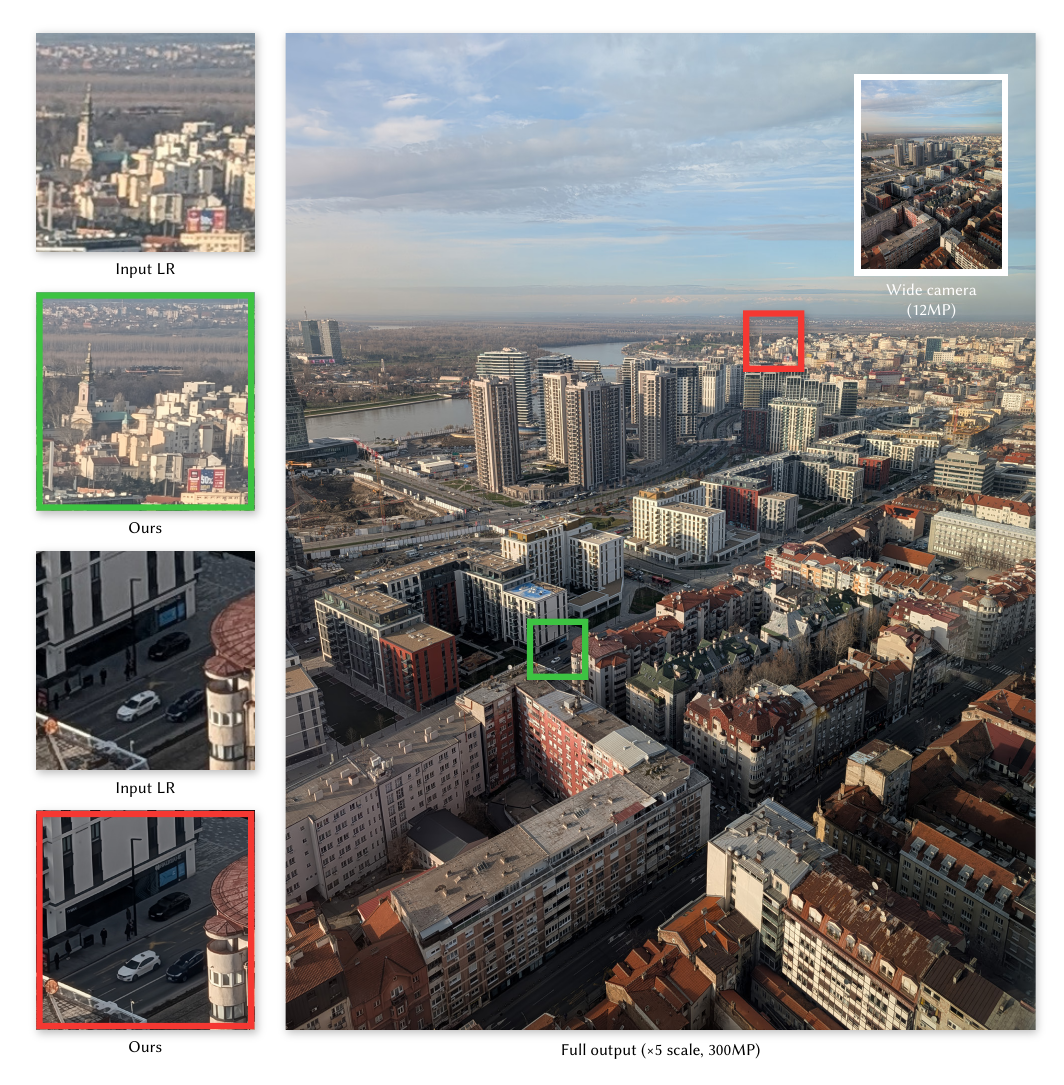}
    \caption{\textbf{Full 300MP reconstruction.} Result produced from a 12MP wide-angle anchor on Google Pixel 10 Pro using 36 telephoto references at $5\times$ zoom. The full $15\mathrm{K}\times20\mathrm{K}$ output is shown together with the wide-angle input inset at its relative scale and two detail crops comparing the LR input with our reconstruction, illustrating recovered fine detail across the FoV. \emph{If this figure does not render correctly, please view the PDF using a standalone viewer.}}
    \Description{A 300MP reconstructed aerial cityscape generated from a 12MP wide-angle image and 36 telephoto references. The original wide-angle image is inset at its relative scale, and two marked locations are enlarged to compare the corresponding low-resolution input crops with the reconstructed crops.}
    \label{fig:fullimage_trimmed}
\end{figure*}

Our system has the following limitations:
1) If references are misaligned, occluded, or invalid due to large motion or disocclusion, the affected regions fall back to LR-conditioned restoration. Without a valid observation, faithful high-frequency detail cannot be recovered, as illustrated by the traffic example in \cref{fig:fig11}.
2) The wide-angle anchor defines the output content and appearance. Clipped highlights, flare, severe underexposure, or very low dynamic range in the anchor cannot be fully corrected from telephoto references without overriding the anchor-defined appearance, as illustrated by the low-light example in \cref{fig:fig11}.
3) The acquisition procedure requires the user to sequentially scan the anchor FoV with multiple telephoto views in a prescribed order, making capture more involved than single-shot or synchronized dual-camera imaging and increasing sensitivity to scene motion over longer acquisitions.
\section{Conclusion}
In this work, we introduce a system for ultra-high-resolution mobile photography. Our guided capture procedure acquires a wide-angle anchor together with a grid of telephoto references covering its FoV; in our experiments, we demonstrate scans with up to $6\times6$ references. The reconstruction pipeline combines cross-view matching and local reference retrieval, transformer-based detail transfer, and generative refinement to produce full-frame outputs at resolutions up to 300MP. We demonstrate the complete pipeline across multiple smartphones with $3\times$--$5\times$ telephoto zoom and show strong quantitative performance of the proposed RefSR component on real dual-camera and $\times4$ RefSR benchmarks. Together, these results demonstrate that multiple telephoto observations can extend high-resolution detail beyond the fixed overlap of conventional dual-camera RefSR while preserving the geometry and appearance defined by the wide-angle anchor.
\begin{acks}
We thank Vladislav Pyatov for helpful discussions and recommendations on image matching, and Nikolai Karetin and Vasiliy Lisovskiy for assisting with scene capture and for lending us their devices, broadening the scope of our testing. We are also grateful to everyone who appeared in our test scenes for their patience during repeated data collection. Finally, we thank the anonymous reviewers for their valuable feedback, which helped improve and strengthen this work.
\end{acks}

\clearpage

\clearpage
\bibliographystyle{ACM-Reference-Format}
\bibliography{main}



\end{document}